%% file: neurips_2025.tex
\documentclass{article}

\usepackage[dblblindworkshop, final]{neurips_2025}

\usepackage[utf8]{inputenc} 
\usepackage[T1]{fontenc}    
\usepackage{hyperref}       
\usepackage{url}            
\usepackage{booktabs}       
\usepackage{amsfonts}       
\usepackage{nicefrac}       
\usepackage{microtype}      
\usepackage{xcolor}         

\usepackage[utf8]{inputenc} 
\usepackage[T1]{fontenc}    
\usepackage{algorithm}
\usepackage{algorithmic}
\usepackage[table,xcdraw]{xcolor}
\definecolor{lightgray}{gray}{0.96}
\definecolor{highlight}{RGB}{224,243,255}
\definecolor{greencheck}{RGB}{0,204,0}
\usepackage{pifont}
\usepackage{tcolorbox}
\usepackage{graphicx}
\usepackage{subcaption}
\usepackage{float}
\usepackage{caption}
\usepackage{newfloat}
\usepackage{listings}
\usepackage{amsmath}
\usepackage{hyperref}
\usepackage{fontawesome5}
\floatstyle{ruled}
\newfloat{listing}{tb}{lst}{}
\floatname{listing}{Listing}

\title{WebGraphEval: Multi-Turn Trajectory Evaluation \\ for Web Agents using Graph Representation}

\author{%
  Yaoyao Qian\thanks{Corresponding author. \texttt{qian.ya@northeastern.edu}} \\
  Northeastern University \\
  \And
  Yuanli Wang \\
  Boston University \\
  \And
  Jinda Zhang \\
  University of Victoria \\
  \And
  Yun Zong \\
  University of Minnesota \\
  \And
  Meixu Chen \\
  Northeastern University \\
  \And
  Hanhan Zhou \\
  George Washington University \\
  \And
  Jindan Huang \\
  Tufts University \\
  \And
  Yifan Zeng \\
  Oregon State University \\
  \And
  Xinyu Hu \\
  University of Texas at San Antonio \\
  \And
  Chan Hee Song \\
  The Ohio State University \\
  \And
  Danqing Zhang \\
  PathOnAI.org \\
}

\begin{document}

\maketitle
\vspace*{-3.2em}
\begin{center}

\vspace{1em}
\href{https://webgrapheval-webpage.vercel.app/}{\faGlobe\ Project Website} \quad
\href{https://web-graph-eval.vercel.app/}{\faLink \ Demo Website}    
\end{center}

\begin{abstract}
Current evaluation of web agents largely reduces to binary success metrics or conformity to a single reference trajectory, ignoring the structural diversity present in benchmark datasets. We present \textbf{WebGraphEval}, a framework that abstracts trajectories from multiple agents into a unified, weighted action graph. This graph representation is directly compatible with existing benchmarks such as WebArena, using both leaderboard trajectories and newly collected runs, and provides a principled basis for analyzing solution spaces without modifying environments. The framework canonically encodes actions, merges recurring behaviors, and applies structural analyses including reward propagation and success-weighted edge statistics. Evaluations across thousands of trajectories from six web agents demonstrate that the graph abstraction captures cross-model regularities, highlights redundancy and inefficiency, and identifies critical decision points overlooked by outcome-based metrics. By framing web interaction as graph-structured data, WebGraphEval establishes a general methodology for multi-path, cross-agent, and efficiency-aware evaluation of web agents.

\end{abstract}

\input{section/1_introduction}
\input{section/2_relatedwork}
\input{section/3_methodology}

\input{section/4_experiments}
\input{section/5_discussion}
\section*{Acknowledgments}{We thank Cookie (Yaoyao’s dog) and Lucas (Yaoyao’s cat) for their comforting presence during this work.}
\bibliographystyle{unsrt}
\bibliography{references}



\input{section/appendix}


\end{document}

%% file: section/1_introduction.tex
\section{Introduction}

Benchmarks such as WebArena \cite{zhou2024webarena} and Mind2Web \cite{deng2023mind2web} provide thousands of recorded trajectories showing how web agents interact with interfaces. These datasets contain many valid solution paths for the same task, but current evaluation methods reduce performance to binary success or to matching a reference trajectory. This leaves out the information in the trajectories themselves, including how agents explore, take detours, or recover.

Existing approaches show this limitation in different forms. Outcome-based metrics ignore the intermediate process. The "LLM-as-a-Judge" paradigm \cite{gu2024survey} allows more flexible judgments but still reduces evaluation to a final outcome. Trajectory conformity \cite{yao2022react,deng2023mind2web} compares against a reference path but reflects the bias of the chosen path, penalizes alternative strategies, and requires frequent updates as interfaces change. In all cases, the common issue is that multiple trajectories are collected, but their overlap, differences, and repeated errors are not used.

We propose that evaluation should use a structured representation of trajectories rather than isolated sequences. Graphs are a natural choice: nodes represent actions, edges represent transitions, and multiple trajectories can be combined into a single graph. This makes it possible to see common strategies, points where agents fail, and how different models behave on the same task.

We present \textbf{WebGraphEval}, a framework that builds such graphs from raw trajectories. The framework converts actions into a canonical form, merges similar behaviors into weighted nodes and edges, and applies analyses such as reward propagation and success-weighted edge classification. Using leaderboard data and additional runs from WebArena, WebGraphEval works with existing benchmarks and can be extended to new agents. Across six web agents, it shows how graph-based evaluation captures shared strategies, redundant actions, and critical steps that outcome-based metrics cannot measure.
This work makes four contributions to web agent evaluation:

\begin{enumerate}
    \item \textbf{WebGraphEval framework.} We present WebGraphEval, a graph-based evaluation framework that aggregates multiple trajectories into a weighted action graph. This representation captures both shared strategies and divergent behaviors, enabling structured analysis that outcome- or conformity-based methods overlook.

    \item \textbf{Cross-agent and benchmark compatibility.} WebGraphEval is applied to leaderboard trajectories and additional runs from WebArena, covering six different agents. This shows that the method works directly with existing benchmarks and enables systematic cross-model comparison.

    \item \textbf{Analytical methods.} The framework includes canonicalization of actions, node merging, and structural analyses such as reward propagation and success-weighted edge statistics. These methods allow us to study task difficulty, agent efficiency, and decision points in a unified way.

    \item \textbf{Practical protocol and tool.} We design an LLM-based annotation protocol to assign necessity labels with 78\% agreement against human judgments, making graph construction scalable. We also provide a visualization interface (\url{https://web-graph-eval.vercel.app/}) for exploring trajectories before and after graph construction.
\end{enumerate}

\begin{table*}[t]
\centering
\caption{Comparison of Web Agent Evaluation Methods}
\label{tab:comparison}
\resizebox{\textwidth}{!}{%
\begin{tabular}{llllll}
\toprule
\textbf{Method/Benchmark}  & \textbf{Graph-based} & \textbf{Multi-path} & \textbf{Analysis Scope} & \textbf{Judge Method} \\
 &  \textbf{Analysis}& \textbf{Support} &  &  \\
\midrule
\rowcolor{lightgray}
WebShop \citep{yao2022webshop} & \textcolor{red}{\ding{55}} & \textcolor{red}{\ding{55}} & Final message (Single) & Rule-based \\
\midrule
Mind2Web \citep{deng2023mind2web} & \textcolor{red}{\ding{55}} & \textcolor{greencheck}{\ding{51}} & Single Trajectory & Rule-based \\
\midrule
\rowcolor{lightgray}
WebArena \citep{zhou2024webarena}  & \textcolor{red}{\ding{55}} & \textcolor{red}{\ding{55}} & Single Trajectory & Rule-based + LLM-as-judge\\
\midrule
VisualWebArena \citep{koh2024visualwebarena}  & \textcolor{red}{\ding{55}} & \textcolor{red}{\ding{55}} & Final message (Single) & Rule-based +  LLM-as-judge \\
\midrule
\rowcolor{lightgray}
WebVoyager \citep{he2024webvoyager} & \textcolor{red}{\ding{55}} & \textcolor{red}{\ding{55}} & Final message (Single) & LLM-as-judge \\
\midrule
Mind2Web 2.0 \citep{gou2025mind2web} & \textcolor{red}{\ding{55}} & \textcolor{red}{\ding{55}} & Single Trajectory & Agent-as-judge \\
\midrule
\rowcolor{lightgray}
AgentBoard \citep{ma2024agentboard} & \textcolor{greencheck}{\ding{51}} & \textcolor{red}{\ding{55}} & Single Trajectory & LLM-as-judge \\
\midrule
ST-WebAgentBench \citep{levy2024stwebagentbench} & \textcolor{red}{\ding{55}} & \textcolor{red}{\ding{55}} & Single Trajectory & Rule-based \\
\midrule
\rowcolor{lightgray}
VideoWebArena \citep{jang2024videowebarena} & \textcolor{red}{\ding{55}} & \textcolor{red}{\ding{55}} & Single Trajectory & Rule-based + LLM-as-judge \\
\midrule
\rowcolor{highlight}
\textbf{WebGraphEval (Ours)}  & \textcolor{greencheck}{\ding{51}} & \textcolor{greencheck}{\ding{51}} & \textbf{Cross Multi Trajectories}  & \textbf{LLM-as-judge} \\
\bottomrule
\end{tabular}%
}
\end{table*}

%% file: section/2_relatedwork.tex
\section{Related Work}

\paragraph{Benchmarks and Evaluation Protocols}  
Benchmarks for web agents range from simulated settings (WebShop \citep{yao2022webshop}) to live websites (Mind2Web \citep{deng2023mind2web}, WebArena \citep{zhou2024webarena}), with extensions to vision, cross-domain, and safety tasks \citep{koh2024visualwebarena,jang2024videowebarena,liu2023agentbench,levy2024stwebagentbench}. Most leaderboards report binary success. Other metrics include progress rate (AgentBoard \citep{ma2024agentboard}), skill scoring (FLASK \citep{ye2024flask}), step-wise conformity to a reference \citep{yao2022react,deng2023mind2web}, and LLM-as-a-Judge outcome checks \citep{gou2025mind2web}. These protocols evaluate each trajectory in isolation and do not analyze the structure across multiple valid paths.

\paragraph{Graph-Based and Structural Approaches}  
Graphs have been applied in many domains to represent sequential decisions and relations. In embodied AI, graphs are used for task planning \citep{shridhar2020alfworld}. In retrieval-augmented generation, systems such as GraphRAG \citep{edge2024local}, LightRAG \citep{guo2024lightrag}, and Causal GraphRAG \citep{haque2025graphrag} build graphs over entities or events to support multi-hop reasoning in document or news QA. Graph-R1 \citep{luo2025graph} uses a hypergraph structure for complex QA tasks. For web agents, WebVoyager \citep{he2024webvoyager} analyzes navigation patterns and Go-Browse \citep{gandhi2025gobrowse} frames exploration as graph search. These works show that graphs can support reasoning and analysis, but they have not been used as a systematic evaluation framework. A side-by-side summary of these approaches is shown in Table~\ref{tab:graph-approaches}. In contrast, WebGraphEval builds directed graphs over canonicalized actions and transitions, aggregates trajectories across agents, and applies graph analysis for evaluation.

%% file: section/3_methodology.tex
\section{Methodology}
\begin{figure}[t]
    \centering
    \includegraphics[width=\linewidth]{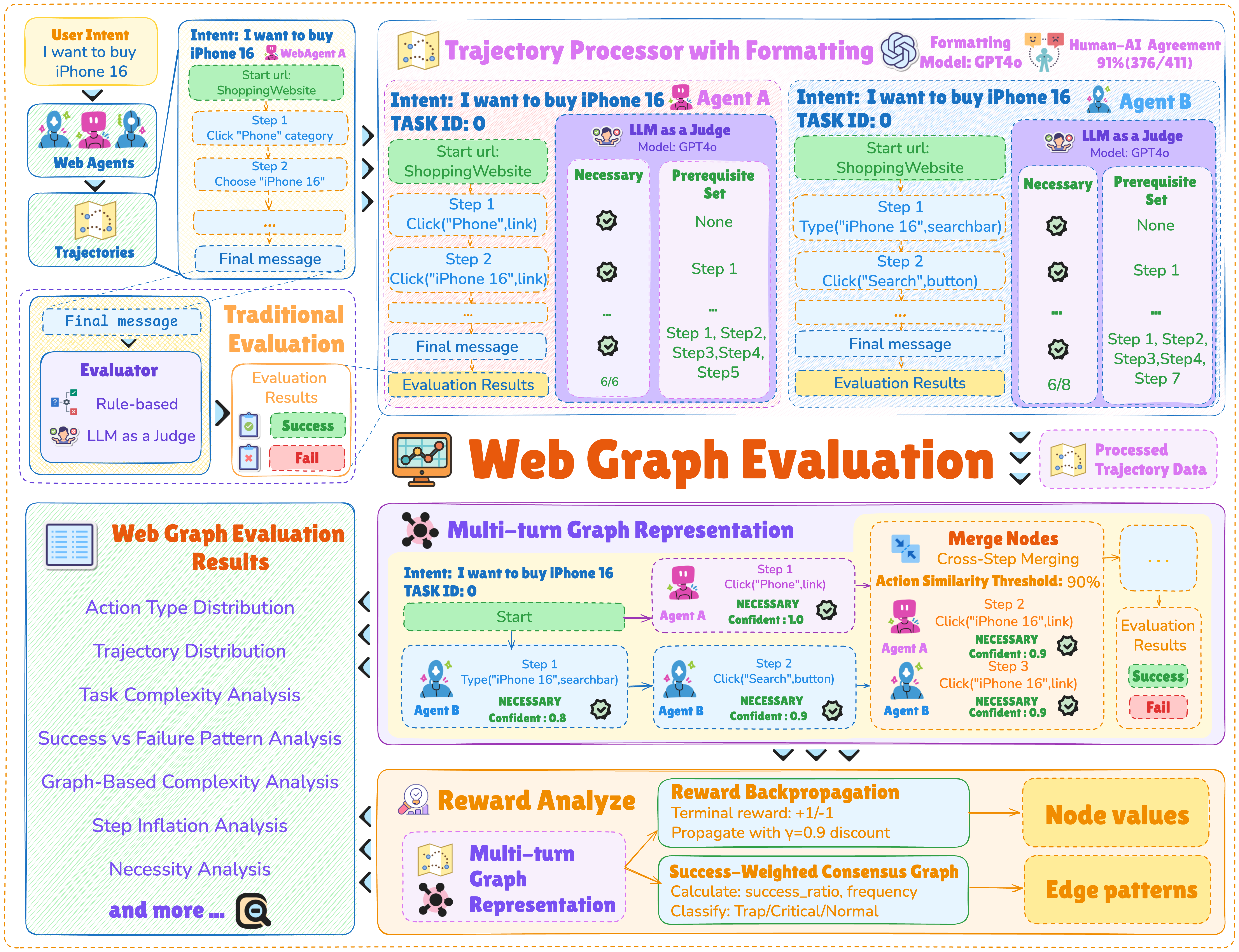}
    \caption{Overview of the WebGraphEval pipeline. Trajectories from multiple agents are first pre-processed and canonicalized into standardized action units. These are merged into a directed action graph, where nodes represent actions and edges represent observed transitions. The graph is then analyzed with reward backpropagation and success-weighted edge classification, and the resulting structure supports multi-dimensional evaluation of agent efficiency, redundancy, and strategy overlap.}
    \label{fig:pipeline}
\end{figure}

WebGraphEval is a graph-based evaluation framework that aggregates trajectories from multiple agents into a unified structure (Figure~\ref{fig:pipeline}). Web interactions are represented as directed transitions between actions: nodes correspond to canonicalized actions, and edges represent observed transitions weighted by frequency and outcome. This abstraction preserves both common strategies and divergent behaviors, enabling evaluation beyond single-path outcomes.  

The framework integrates two complementary analysis methods. Reward backpropagation propagates task outcomes backward through trajectories to estimate the value of earlier actions, while success-weighted edge classification labels transitions by their empirical association with success or failure. These provide temporal and structural perspectives on agent behavior.  

As illustrated in Figure~\ref{fig:pipeline}, the pipeline unfolds in four stages. Trajectories are first pre-processed and canonicalized into standardized actions. Canonical actions are then merged into nodes, and transitions are stored as directed edges. The resulting graph is analyzed with reward backpropagation and edge classification. Finally, the graph supports multi-dimensional evaluation of agents, including efficiency, redundancy, and strategy comparison across models.

\subsection{Trajectory Formalism and Pre-processing}

The first stage of WebGraphEval is to convert raw agent trajectories into a formal representation suitable for analysis. Each trajectory $\tau = (a_1, \ldots, a_n)$ is an ordered sequence of actions, where each action $a_i = \langle \text{desc}_i, \text{url}_i \rangle$ consists of a textual description and the corresponding execution URL. A key challenge is that agents often describe the same action in different natural language forms. To address this, we use an LLM-based canonicalization step that maps diverse descriptions to a standardized schema. For example, ``clicked on the `Submit' button'' and ``press Submit'' are both converted to the unified form \texttt{click(text=`Submit', element=`button')}.

Once actions are normalized, trajectories are labeled as successful or failed using an LLM-as-Judge protocol. The model is prompted with task-specific rules and evaluates whether the final state of each trajectory satisfies the original task requirements. This step produces two disjoint sets: successful trajectories $D_{\text{success}}$ and failed trajectories $D_{\text{fail}}$.

Finally, we refine the representation by identifying redundant actions. Each action $a_i$ is assigned a necessity label $\nu(a_i) \in \{0,1\}$, where $\nu(a_i)=1$ indicates that the action is essential for task completion. This annotation is generated automatically by an LLM and spot-checked with human evaluators to confirm reliability. Figures~\ref{fig:prompt-action-conversion-system} and \ref{fig:prompt-action-conversion-user} show the prompts used for action canonicalization and necessity labeling.
\subsection{Consensus Graph Construction}

Given canonicalized trajectories, we construct a consensus directed graph $G=(V,E)$ that serves as the representation for subsequent analysis. The input is a set of action sequences $\tau=(a_1,\ldots,a_n)$ from multiple agents, and the output is a graph in which each node $v\in V$ corresponds to an equivalence class of actions while each edge $(u\!\to\!v)\in E$ represents an observed transition. Nodes and edges carry counts and outcome-conditioned statistics that are used in later stages of evaluation.

To create the node set, we merge semantically similar actions into common nodes. Each action is already expressed in a structured format after canonicalization, such as \texttt{click(text=\dots, element=\dots)}. We measure the similarity between two actions $a_i$ and $a_j$ with a normalized edit distance,
\[
\mathrm{sim}(a_i,a_j)=1-\frac{\mathrm{Levenshtein}(a_i,a_j)}{\max(|a_i|,|a_j|)}\,.
\]
Merging is applied in two passes. In step-based merging, actions are only compared with others that occur at the same time index across trajectories, which preserves temporal roles. In cross-step merging, comparisons extend across different positions, which allows the graph to capture actions that recur in multiple places. The merging process is implemented with a deterministic union–find procedure to avoid dependence on processing order.

The similarity threshold $\theta$ is a tunable parameter. A high value leaves near-duplicates unmerged, producing fragmented graphs, while a low value risks collapsing distinct actions. We use $\theta=0.9$ as a practical compromise: it removes superficial wording differences while still maintaining distinctions between genuinely different behaviors. This threshold works without task-specific tuning, although future work could replace it with embedding-based similarity or learned clustering.

Edges are then created from adjacent actions in each trajectory. For every pair $(a_t,a_{t+1})$, we add a directed edge $(\pi(a_t)\!\to\!\pi(a_{t+1}))$, where $\pi$ maps actions to their merged node. Each edge stores the number of times it occurs, its counts under successful and failed trajectories, and the empirical success rate derived from these counts. Nodes also retain step-index histograms to support later analysis of temporal patterns. The pairwise similarity computation has quadratic complexity in the number of actions. Since our dataset is moderate in size, we perform the comparisons directly. For larger datasets, approximate blocking strategies (e.g., by action type or host) could reduce the cost to near-linear, but we leave this as future work.

\subsection{Dual Reward Mechanisms}

The consensus graph aggregates trajectories from all six agents, but to make sense of this structure we need a way to measure which actions and transitions matter most. WebGraphEval employs two complementary reward mechanisms. The first propagates outcome signals along entire trajectories, highlighting how early decisions contribute to eventual success or failure. The second examines transition statistics directly, identifying structural patterns such as traps, bottlenecks, and critical paths. Together these methods turn raw trajectory data into interpretable evidence about agent behavior.

The temporal reward backpropagation mechanism is inspired by reinforcement learning. Each terminal state is initialized with a reward of $+1$ if the task succeeds and $-1$ if it fails. These values are then propagated backward through the graph so that earlier nodes inherit credit or blame from their successors. Formally, the value $V(v)$ of a node $v$ is defined in terms of its outgoing edges: each successor contributes proportionally to how often the transition is observed, discounted by a factor $\gamma=0.9$ to reduce the influence of distant outcomes. This procedure assigns positive values to actions that reliably lead, through several steps, to task completion, and negative values to those that tend to end in failure. In this way, the graph reveals decision points that matter even if their effects only appear later in the trajectory.

The success-weighted consensus analysis takes a different perspective. Instead of propagating signals through paths, it looks at the observed frequency and success ratio of each edge. For an edge $e$, the success ratio $s(e)$ is the proportion of times it appears in successful trajectories, and the weight $w(e)$ is its relative frequency compared to all edges. Using these two measures, edges are classified into interpretable categories. Trap edges are those that occur often but almost always lead to failure, such as clicking a misleading link. Critical edges are rare but consistently successful transitions, representing expert-like behavior. Bottleneck edges are high-frequency transitions with mixed success, often corresponding to fragile but necessary steps like form submissions. All other edges are treated as normal, reflecting routine navigation without strong outcome bias. This classification exposes how agents collectively approach tasks and where they are most prone to error.

Beyond individual edges, we also estimate the importance of each node. A node is considered important if it is both frequently visited and strongly associated with successful outcomes. We capture this with a score that averages the success ratios of incoming and outgoing edges and multiplies by the node’s visitation frequency. Intuitively, this highlights actions that not only appear often in trajectories but also play a decisive role in task completion. An example would be reaching a login page, which is both common and essential for downstream success.
\subsection{Multi-dimensional Behavioral and Cross-Agent Evaluation}

The final stage of WebGraphEval evaluates agent behavior across both individual and comparative dimensions. At the trajectory level, we assess path optimality by comparing observed trajectory lengths with the shortest successful paths in the dataset, excluding anomalous cases. This metric captures inefficiency arising from unnecessary exploration. Leveraging LLM-based annotations, we further distinguish necessary from redundant actions, providing a fine-grained measure of how directly agents pursue task goals. Temporal dynamics are also incorporated by analyzing how action types are distributed across early, middle, and late phases of a trajectory, revealing when redundancy or recovery attempts are most likely to occur.

These trajectory-level measurements are then aggregated into agent-level profiles. Beyond raw success rates, we examine performance across task categories, complexity levels, and trajectory lengths, which highlight characteristic strategy patterns such as average path length, preferred navigation choices, and resilience to increasing task difficulty. To extend the analysis further, WebGraphEval enables cross-agent comparisons on shared tasks. Using entropy-based metrics and clustering over trajectory distributions, we quantify the diversity of strategies, determining whether models converge on similar behaviors or adopt distinct approaches. The consensus graph also identifies action sequences that consistently appear in successful trajectories across multiple frameworks, pointing to robust strategies that generalize beyond individual agents.

%% file: section/4_experiments.tex
\section{Experiments}

\subsection{Dataset and Setup}

Our experiments are conducted on a new trajectory dataset built from the WebArena benchmark \cite{zhou2024webarena}. The dataset consists of 4,768 trajectories collected from six agent frameworks attempting 812 unique tasks. The resulting corpus produces large and diverse action graphs, with 40,431 nodes and 45,656 edges in total, averaging 49.79 nodes and 56.23 edges per task. Across all trajectories, we observe 40,888 individual actions, including 19,380 clicks (47.4\%), 8,312 type actions (20.3\%), and 1,302 select operations (3.2\%). Overall, 2,180 trajectories were successful (45.7\%) and 2,588 failed (54.3\%), providing balanced coverage of both effective and ineffective strategies.

All trajectories were evaluated consistently using an LLM-based judge (\texttt{o4-mini-2025-04-16}, temperature 0.1) to determine success or failure, and action-level necessity labels were obtained with \texttt{GPT-4o-2024-08-06}. Confidence scores assigned by the LLM are consistently high across frameworks (0.926–0.976), indicating stable judgments independent of outcome. To validate annotation quality, each framework was evaluated with three independent runs, and human evaluation was performed on sampled subsets. Canonicalization achieved 91\% agreement with human annotators (376/411), while necessity judgments reached 78\% agreement (404/520). These checks confirm that LLM-based annotations are sufficiently reliable for downstream analysis.

For graph construction, actions were merged using a normalized edit similarity threshold of $\theta=0.9$. This threshold reduces superficial variation across agent implementations while preserving distinct behaviors, and ensures that common strategies are captured without erasing meaningful diversity. The resulting graphs reveal that 87.4\% of tasks involve at least one merge, though only 5.6\% of nodes are merged overall, highlighting both structural overlap and significant diversity in agent behavior.

Finally, two structural characteristics of the dataset are noteworthy. First, a subset of tasks (91 cases, about 13\% of all successes) contain one-step successful trajectories, which likely reflect differences in action granularity or pre-filled states; these anomalies are retained for completeness but analyzed separately later. Second, outcome agreement across frameworks is limited: among 761 tasks attempted by all six agents, only 29 tasks (3.8\%) were solved universally, 99 tasks (13.0\%) failed universally, and the majority (83.2\%) showed mixed outcomes. This heterogeneity underscores the need for evaluation methods that integrate across diverse agent behaviors rather than assuming a single reference trajectory.

\subsection{Framework-Level Statistics}
\begin{table*}
  \centering
  \caption{Framework-level performance comparison evaluated using
  \texttt{llm\_judge} (single evaluation).}
  \label{tab:framework-performance-extended}
  \resizebox{\textwidth}{!}{%
  \begin{tabular}{@{}lcccccc@{}}
  \toprule
  \textbf{Framework} & \textbf{Success} & \textbf{Failure} &
  \textbf{Success Rate} & \textbf{Avg Steps} & \textbf{Avg.
  Confidence} & \textbf{Necessity Rate} \\
  \midrule
  \textbf{Zeta Labs Jace.AI} \cite{jaceai2025} & 526 & 286 & 64.78\% &
   5.6 & 0.934 & 73.7\% \\
  \textbf{IBM CUGA} \cite{ibm-cuga-2024} & 477 & 331 & 59.03\% & 5.3 &
   0.972 & 80.6\% \\
  \textbf{Learn by Interact} \cite{learnbyinteract} & 440 & 372 &
  54.19\% & 6.0 & 0.967 & 72.9\% \\
  \textbf{UI-TARS} \cite{uitars} & 296 & 468 & 38.74\% & 13.2 & 0.976
  & 82.0\% \\
  \textbf{OpenAI-CUA} \cite{openai-cua-2025} & 226 & 582 & 27.97\% &
  6.4 & 0.932 & 74.1\% \\
  \textbf{BrowserUse} \cite{browseruse2025} & 215 & 549 & 28.14\% &
  15.6 & 0.926 & 74.7\% \\
  \midrule
  \textbf{Total} & \textbf{2,180} & \textbf{2,588} & \textbf{45.75\%}
  & -- & -- & -- \\
  \bottomrule
  \end{tabular}%
  }
  \end{table*}

We first compare performance across the six agent frameworks. Table~\ref{tab:framework-performance-extended} reports success and failure counts, success rates, average trajectory lengths, confidence scores, and necessity rates. Success rates vary widely, from 64.78\% for Jace.AI—the strongest overall performer—to 27.97\% for OpenAI-CUA. Trajectory lengths also differ: IBM CUGA completes tasks in an average of 5.3 steps, while BrowserUse requires 15.6. Necessity rates range from 72.9\% to 82.0\%, reflecting differences in how directly agents pursue task goals. The uniformly high confidence (0.926–0.976) indicates that the LLM-as-judge provides stable and reliable evaluations, ensuring that performance differences reflect true behavioral variation rather than judging noise.

\begin{figure}[t]
  \centering
  \includegraphics[width=0.55\textwidth]{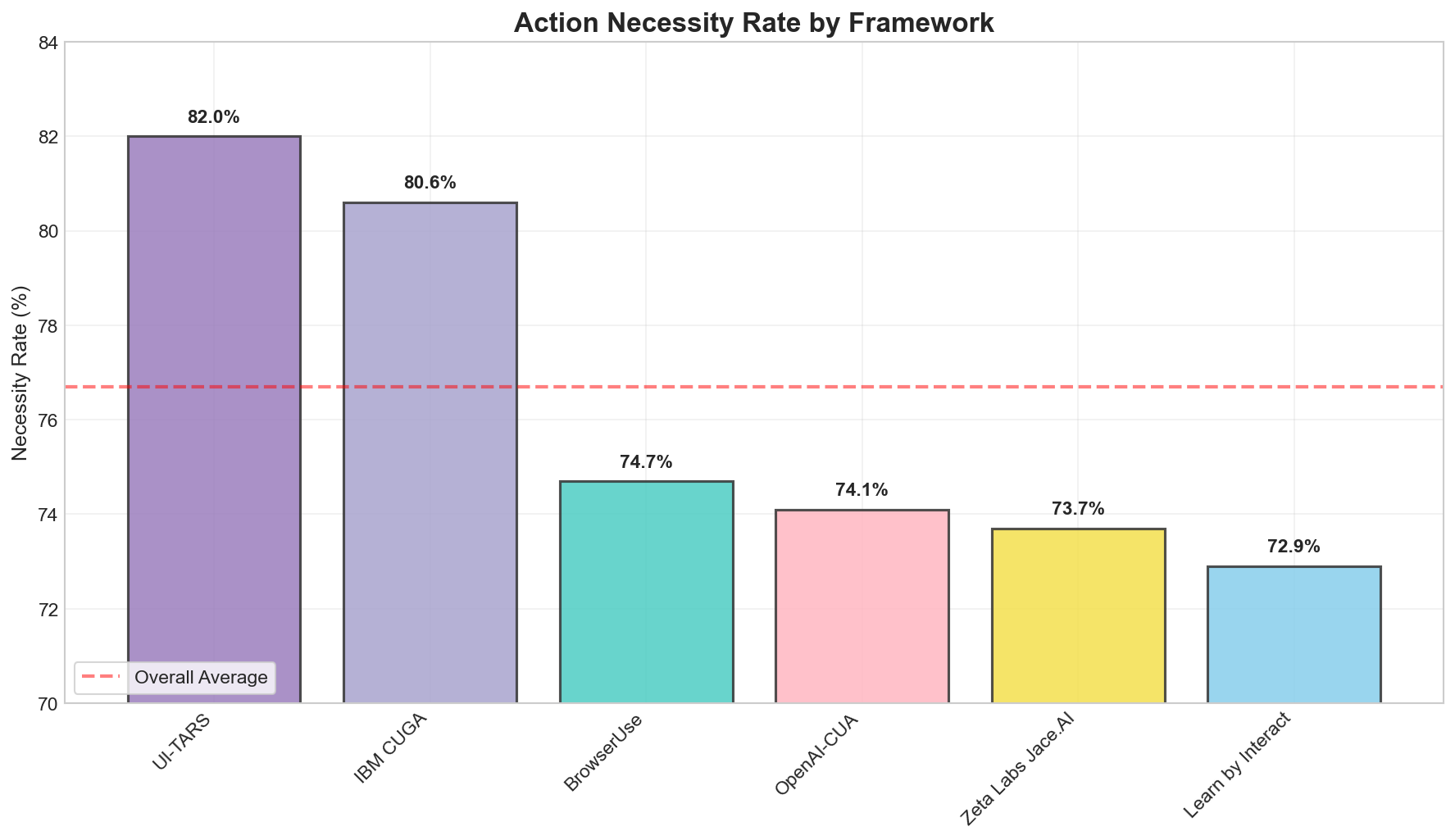}
  \includegraphics[width=0.42\textwidth]{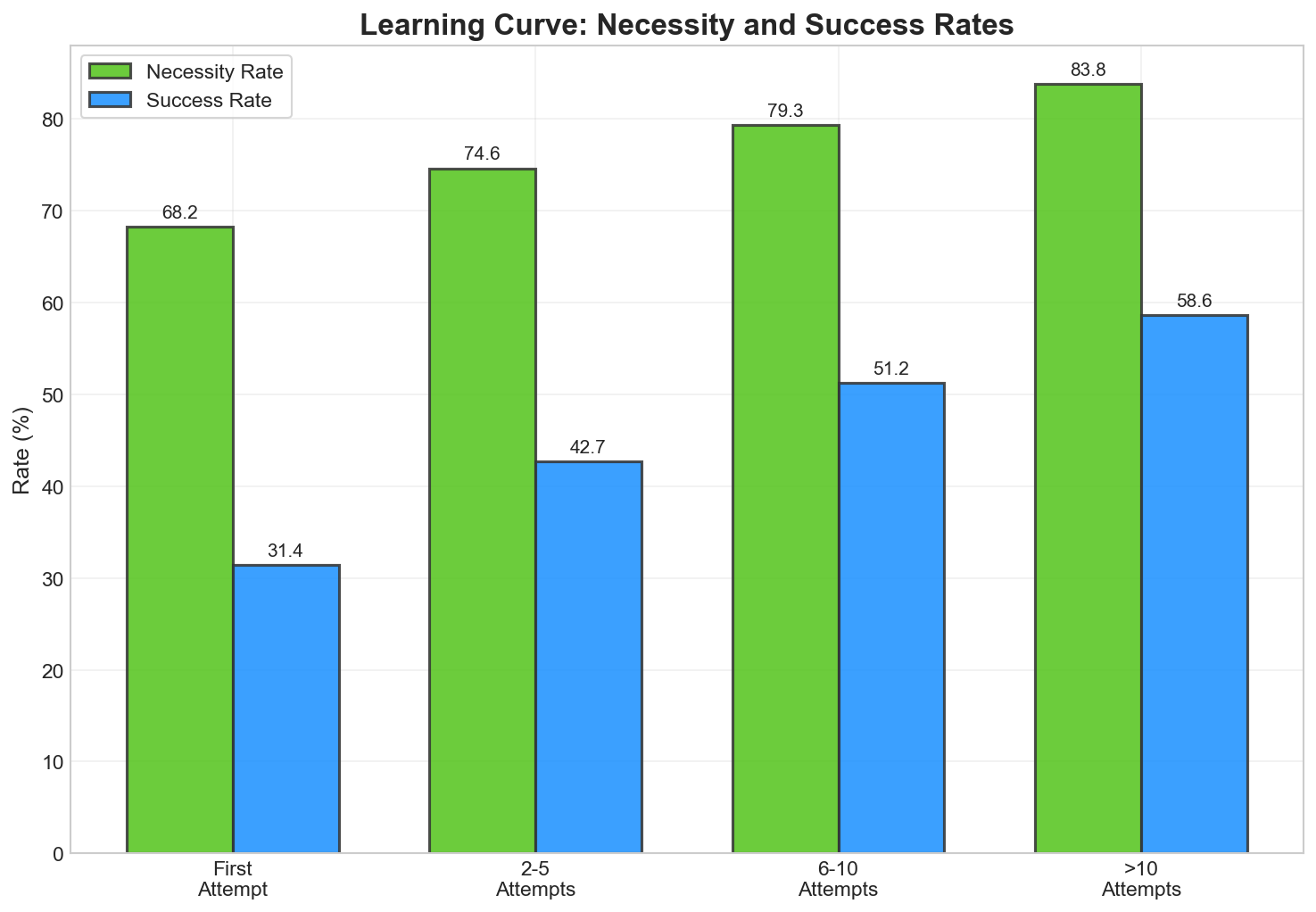}
  \caption{Action necessity across frameworks and learning dynamics. 
  (Left) Necessity rates of six frameworks, showing variation from 72.9\% to 82.0\%. 
  (Right) Learning curve showing how necessity rates improve with repeated task attempts, 
  suggesting necessity is a learnable signal for agent optimization.}
  \label{fig:necessity_framework_learning}
\end{figure}

These results demonstrate that efficiency and effectiveness are not tightly coupled. Both UI-TARS and IBM CUGA maintain focused action sequences, as indicated by high necessity rates, yet their success rates differ substantially (38.7\% vs.\ 59.0\%). This mismatch suggests that minimizing redundancy alone is insufficient: agents must also make correct decisions at critical points. Figure~\ref{fig:necessity_framework_learning} (left) illustrates this divergence, showing that high necessity does not always translate into high success.

Necessity also evolves with experience. Figure~\ref{fig:necessity_framework_learning} (right) shows a learning curve where necessity rises from 68\% on first attempts to over 83\% after ten. This trend indicates that necessity is a learnable signal: with repeated exposure, agents reduce redundant actions and become more efficient.

Two structural characteristics further qualify these results. First, 91 tasks (13\% of all successful cases) contain one-step successes, likely due to action granularity or pre-filled states. These anomalies are retained but analyzed separately to avoid distorting efficiency metrics. Second, cross-framework agreement is limited: of 761 tasks attempted by all six agents, only 29 (3.8\%) were solved universally and 99 (13.0\%) failed universally, while the majority (83.2\%) showed mixed outcomes. This heterogeneity underscores that no single framework dominates and highlights the value of integrating across diverse agent behaviors rather than relying on a single reference trajectory.

\subsection{Graph-Level Structural Analysis}

Consensus graphs reveal clear structural differences across agents. IBM CUGA and Learn by Interact, with shorter average trajectories, generate compact and linear graph topologies, whereas UI-TARS and BrowserUse produce highly exploratory graphs with multiple branches. This divergence reflects different balances between goal-directed execution and exploratory search.

\begin{figure}[t]
  \centering
  \includegraphics[width=0.48\textwidth]{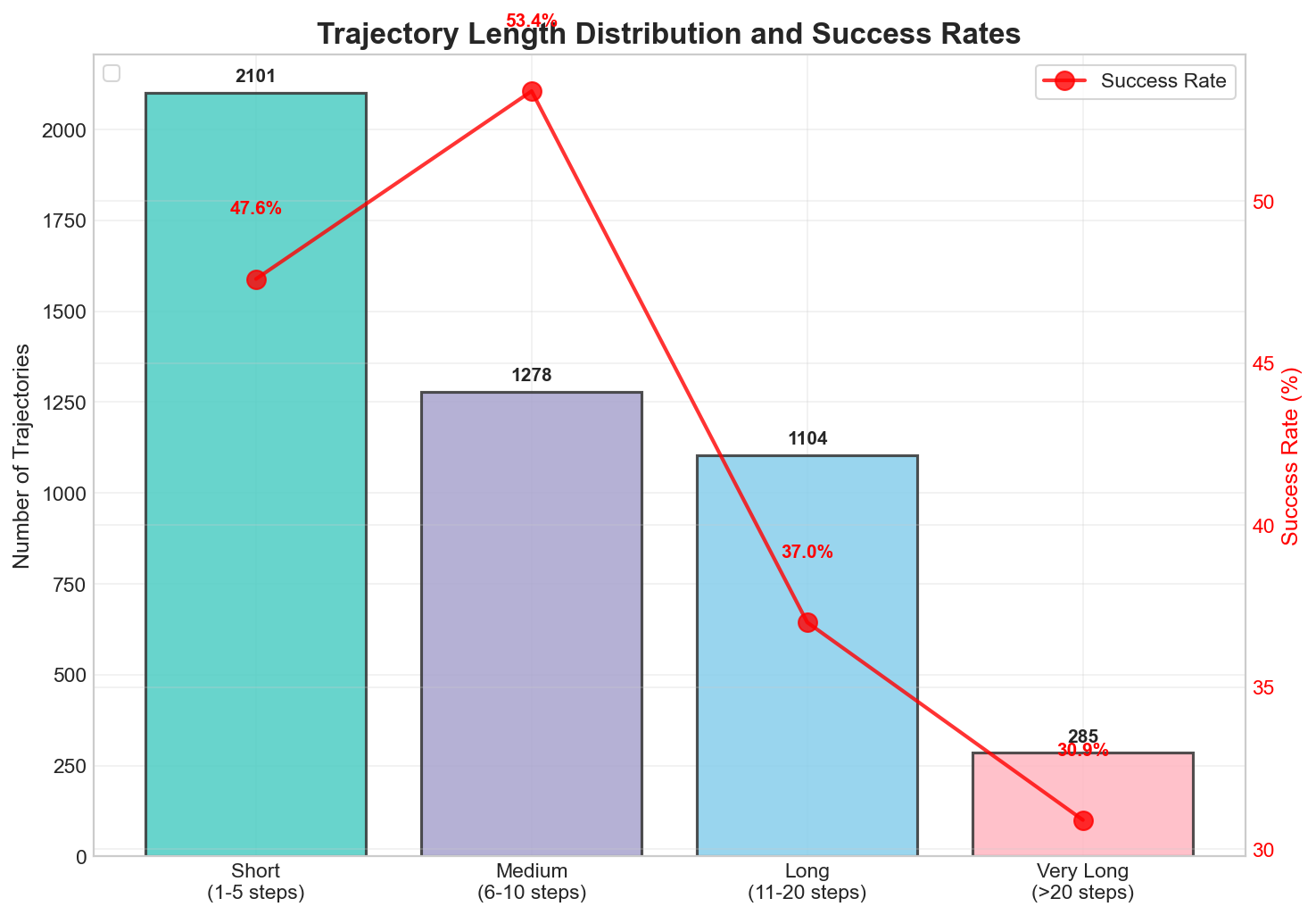}
  \includegraphics[width=0.48\textwidth]{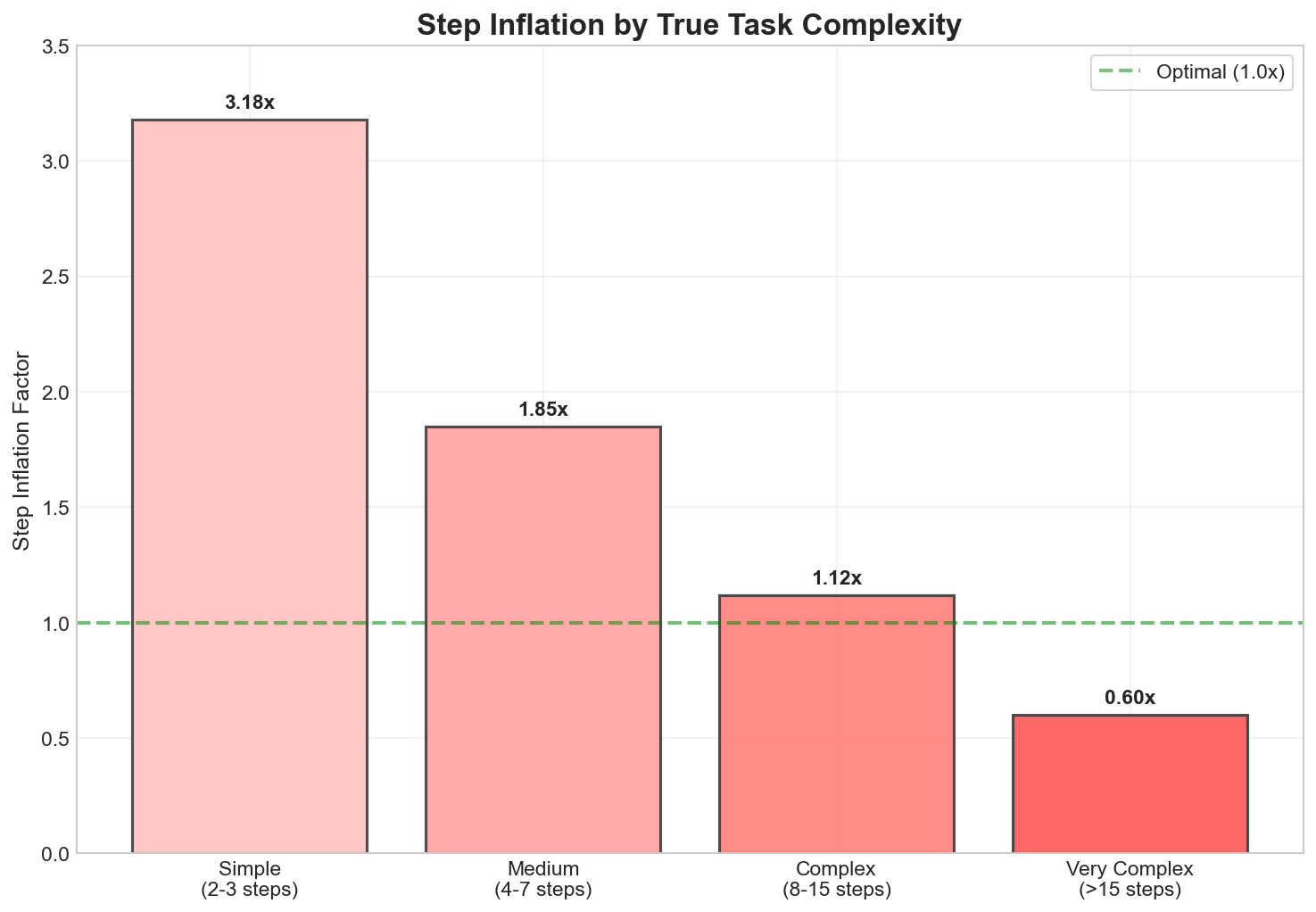}
\caption{Trajectory efficiency and success patterns. 
(Left) Success rates follow an inverted-U curve with respect to trajectory length, peaking at medium trajectories (6–10 steps, 53.4\%). 
(Right) Step inflation by task complexity shows a counter-intuitive trend: agents inflate simple tasks the most (3.18$\times$), while complex tasks approach or even beat near-optimal paths due to survival bias.}

  \label{fig:graph_complexity_success}
\end{figure}

Task success follows a non-linear pattern with respect to complexity: simple tasks succeed 44.1\%, medium-complexity tasks peak at 54.0\%, and performance declines to 45.3\% and 31.0\% for complex and very complex tasks. We formalize complexity as
\[
\text{Complexity} = \frac{\text{nodes} \times \text{edges}}{\text{trajectories}},
\]
and observe a 7.9\% degradation from low- to high-complexity tasks (Figure~\ref{fig:graph_complexity_success}), confirming structural complexity as a meaningful predictor of difficulty.

Efficiency is further captured by step inflation, the ratio between observed steps and the shortest successful path. The average inflation is 2.14$\times$, with some agents taking over 15 steps for 4-step solvable tasks. Frameworks with higher structural efficiency, such as IBM CUGA (80.6\% necessity rate), achieve stronger success compared to less focused agents like Learn by Interact (72.9\%), showing that necessity and inflation jointly capture key behavioral differences.

As shown in Figure~\ref{fig:graph_complexity_success} (left), trajectory length and success rates exhibit a clear inverted-U pattern: medium-length trajectories (6–10 steps) achieve the highest success rate (53.4\%), while both short (47.6\%) and very long trajectories (30.9\%) underperform. This indicates that tasks of moderate length provide the right balance between feasibility and planning opportunity, whereas trivial or overly long trajectories expose limitations in current agents.

Figure~\ref{fig:graph_complexity_success} (right) analyzes step inflation across complexity levels, defined as the ratio between an agent’s actual steps and the shortest successful path. Surprisingly, simple tasks show the largest inflation (3.18$\times$), suggesting agents frequently take unnecessarily long detours even when optimal solutions are short. In contrast, complex and very complex tasks display lower inflation (1.12$\times$ and 0.60$\times$), likely because only highly efficient trajectories survive, creating a selection bias toward near-optimal solutions. Together, these results show that both trajectory length and inflation provide complementary signals about structural efficiency.

\subsection{Action Necessity Analysis}

Overall, 76.7\% of actions are labeled necessary by our LLM-based annotation system, showing that most behaviors directly contribute to task completion while a substantial 23.3\% are exploratory or redundant. Framework-level differences are modest but notable: necessity rates range from 72.9\% (Learn by Interact) to 82.0\% (UI-TARS). Importantly, high necessity alone does not guarantee success—UI-TARS has the highest necessity rate but only a 38.7\% success rate—indicating that efficient action selection must be paired with correct decision-making at key points.

Necessity varies systematically by action type. \texttt{Type} actions are most often essential (82.0\%), followed by \texttt{Select} (79.7\%) and \texttt{Click} (78.2\%), while “other” actions are least necessary (70.2\%). This suggests that data-entry operations nearly always advance task goals, whereas miscellaneous navigations are more prone to redundancy.  

Task complexity and action confidence provide additional perspectives. As tasks grow more complex, the proportion of unnecessary actions rises steadily, from 15.8\% in simple tasks to 31.1\% in very complex tasks. Confidence strongly predicts necessity: high-confidence actions ($>$0.95) are necessary in 83.4\% of cases, compared to only 61.7\% for low-confidence actions ($<$0.85), with correlation $r=0.67$, $p<0.001$. Temporally, early actions are most critical (84.3\% necessity in the first three steps), while later actions drop to 71.2\%, reflecting the accumulation of detours and recovery attempts in longer trajectories.  

Necessity is also dynamic rather than fixed. Figure~\ref{fig:necessity_framework_learning} (right) shows that necessity rates increase with repeated task attempts, rising from 68\% on first attempts to over 83\% after ten. This learning curve demonstrates that agents can improve efficiency over time by reducing redundant actions. At the same time, anomalies such as one-step successes (13\% of all successful trajectories) show that necessity can be misleading in edge cases, since some tasks can be completed with minimal effort due to action granularity or pre-filled states. Together, these results establish necessity as a measurable, learnable, and interpretable signal for evaluating agent efficiency, while also highlighting its limitations.

\subsection{Behavioral Analysis}

Consensus graphs provide a lens for understanding navigation strategies at both the path and action levels. At the trajectory scale, we observe three dominant styles of behavior. Direct navigation strategies follow near-optimal paths with minimal detours and account for 43\% of successful trajectories. Exploratory navigation strategies, seen in 31\% of cases, involve backtracking and alternative path trials before arriving at the goal. The remaining 26\% of trajectories exhibit hybrid approaches, where targeted actions are interleaved with selective exploration. This spectrum of strategies highlights that agents vary not only in whether they succeed but also in how they approach problem solving.

At the action scale, necessity analysis shows that 76.7\% of actions are essential for task completion. Clear differences emerge across action types: \texttt{Type} actions are the most consistently necessary (82.0\%), followed by \texttt{Select} (79.7\%) and \texttt{Click} (78.2\%), while miscellaneous actions are the least necessary (70.2\%). Correlation with LLM confidence ($r=0.68$) further indicates that redundant or exploratory actions are often accompanied by lower certainty. Temporally, earlier steps carry disproportionate importance, with 84.3\% necessity in the first three actions versus only 71.2\% after step ten. Longer trajectories therefore accumulate more redundancy and recovery attempts, making necessity a meaningful signal for distinguishing efficient from inefficient behaviors.
\subsection{Cross-Agent Analysis}

Moving from individual trajectories to collective behavior, we analyze cross-agent variation using entropy-based diversity metrics and clustering over trajectory distributions. Strikingly, 83.2\% of tasks produce mixed outcomes across agents, underscoring that different frameworks bring complementary strengths rather than converging on a single strategy. Zeta Labs Jace.AI achieves its strongest performance on content creation tasks (66\%), while IBM CUGA leads in structured updates (61\%). Other agents reveal their own niches, reinforcing the view that no single framework dominates across task categories.

Consensus graphs further reveal shared structural backbones. Across models, 89\% of successful trajectories begin with similar initial sequences, suggesting the presence of critical paths that anchor successful behavior. At the same time, 37\% of failed trajectories terminate prematurely compared to their successful counterparts. This pattern implies that agents often recognize futility early, abandoning tasks after encountering traps rather than engaging in extended redundant exploration.

\subsection{Key Findings}

Our analyses yield several insights into web agent behavior. First, we observe a performance–efficiency trade-off: higher necessity rates signal more focused action sequences but do not always correlate with higher success, as illustrated by UI-TARS. Second, task complexity shows an inverted-U relationship with success, with medium-complexity tasks achieving the highest rates while both simple and very complex tasks prove more difficult. Third, frameworks exhibit clear complementarity, with different agents excelling in distinct categories and the majority of tasks showing heterogeneous outcomes across models. Finally, consensus graphs capture behavioral phenomena that single-path metrics overlook, such as the existence of shared critical paths and the tendency of failed trajectories to terminate early. Together, these findings demonstrate that WebGraphEval not only benchmarks performance but also surfaces the structural and behavioral dynamics that define agent strengths and weaknesses.

%% file: section/5_discussion.tex
\section{Discussion} In this work, we introduced \textbf{WebGraphEval}, a framework that extends evaluation beyond binary success rates by providing a structured, interpretable, and multi-dimensional view of web agent behavior. The results demonstrate that WebGraphEval captures not only whether agents succeed, but also how they navigate, where inefficiencies occur, and which strategies are shared or divergent across frameworks. This enables a richer understanding of strengths, weaknesses, and complementarities in current web navigation systems. 

Despite these contributions, several limitations remain. First, the reliability of consensus graphs depends on the availability of diverse trajectories. Tasks with few attempts or limited agent coverage yield less stable structural insights. Second, the current state and action canonicalization is implemented through heuristics and LLM-based prompts. While effective in many cases, this approach may struggle with the breadth of real-world interfaces and actions. Third, contextual completeness is constrained by the dataset: many trajectories lack full screenshots or auxiliary information, which limits the fidelity of environment reconstruction. 

Looking forward, we see three main avenues for improvement. (1) \textbf{Reducing data dependence}: Few-shot graph construction and transfer learning across related tasks can extend the framework’s applicability to sparse or novel domains. (2) \textbf{Improving abstraction}: Replacing heuristic canonicalization with learned, semantically informed models can yield more robust state and action representations. (3) \textbf{Closing the loop}: Rather than being purely diagnostic, consensus graphs could inform online decision-making—either by guiding agent exploration during inference or by serving as a structured reward signal in reinforcement learning.

%% file: section/appendix.tex
\clearpage
\appendix
\onecolumn

\section*{Appendix}
\label{sec:appendix}

\appendix
\section{Supplementary Figures}

This appendix presents additional analyses and visualizations that complement the main text. To avoid redundancy, we organize related figures into compact multi-panel layouts.

\subsection{Action Necessity Analysis}

\begin{figure}[ht]
\centering
\begin{subfigure}[b]{0.48\textwidth}
    \centering
    \includegraphics[width=\textwidth]{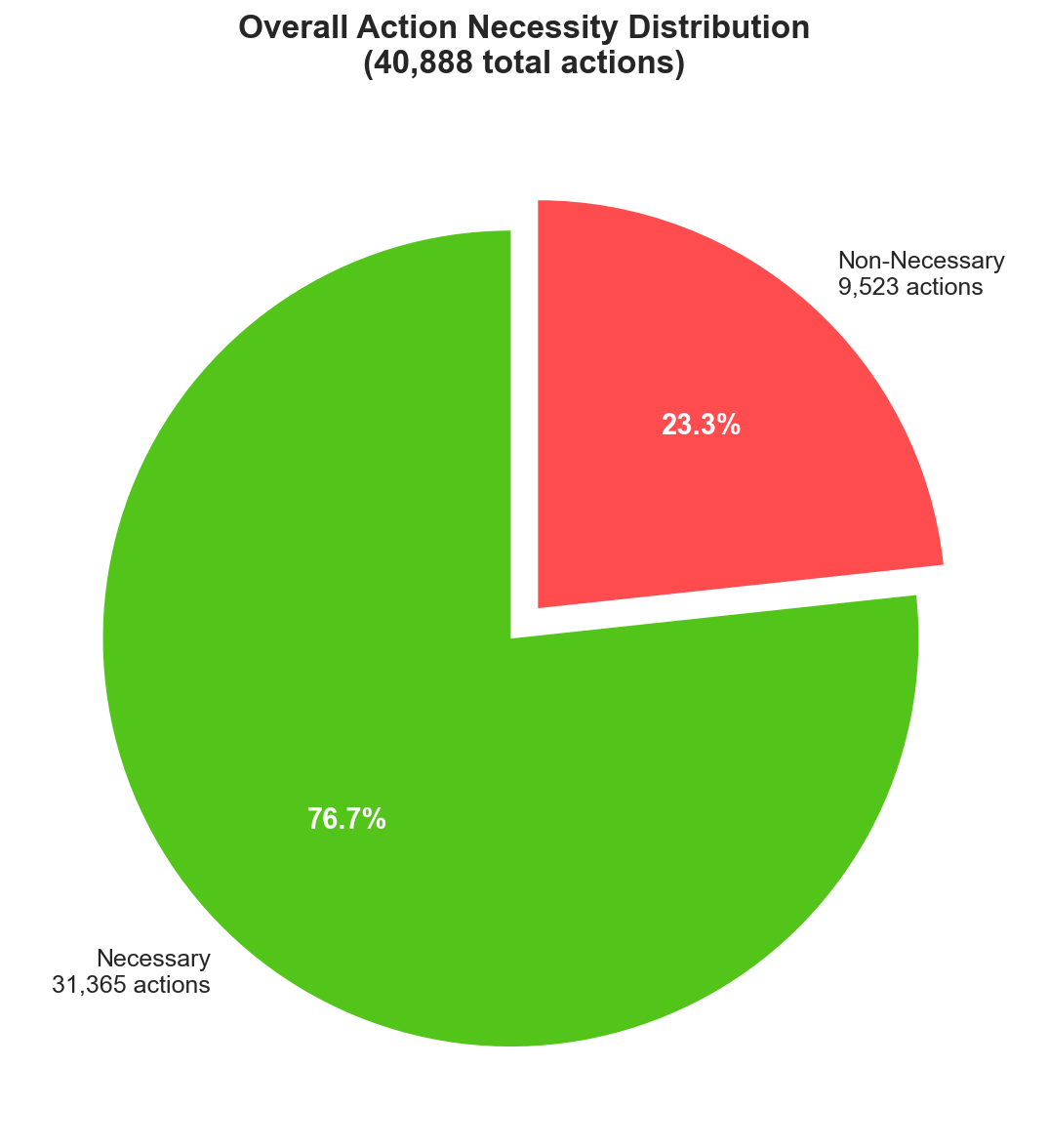}
    \caption{Overall necessity distribution (76.7\% of 40,888 actions are necessary).}
\end{subfigure}
\hfill
\begin{subfigure}[b]{0.48\textwidth}
    \centering
    \includegraphics[width=\textwidth]{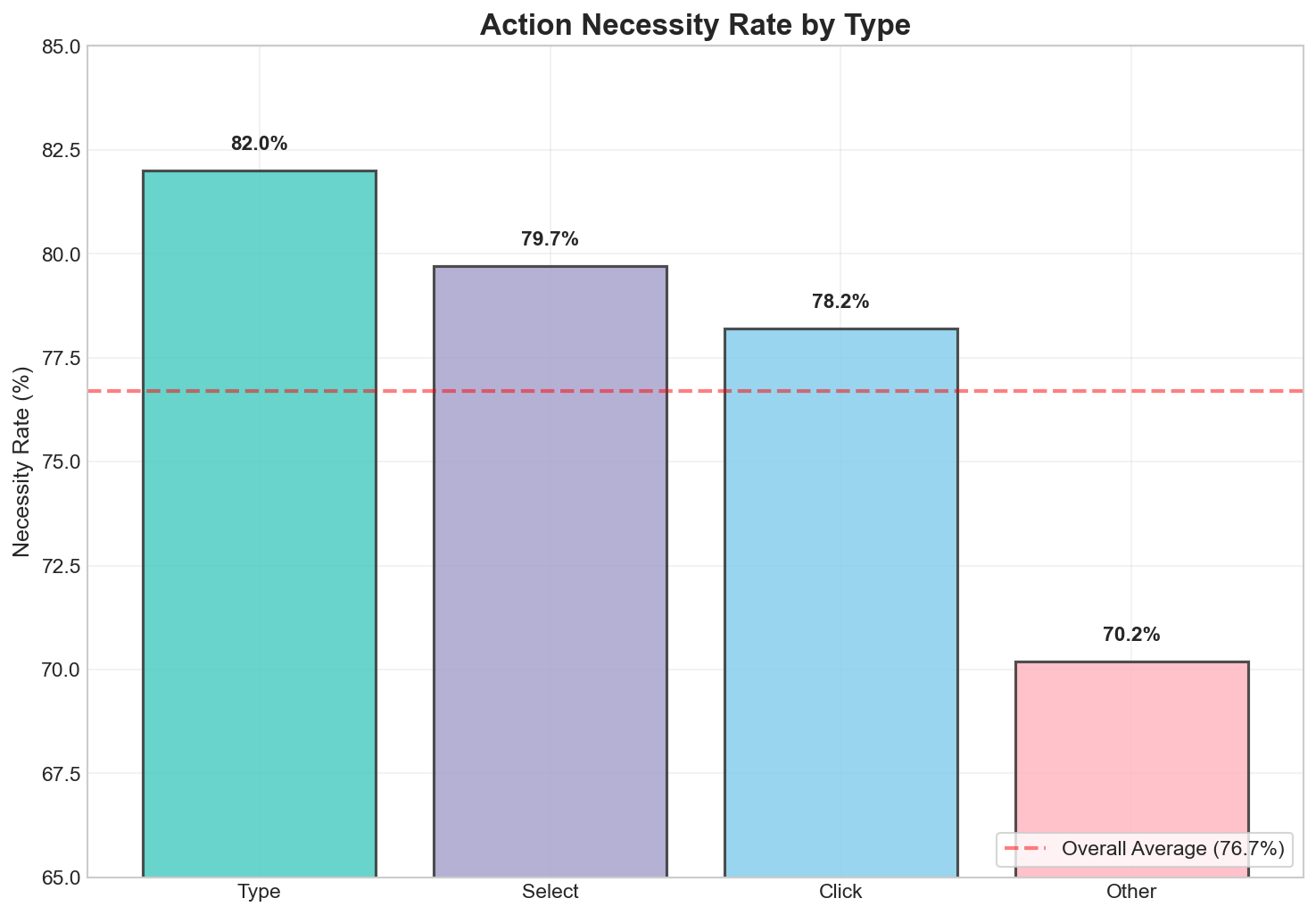}
    \caption{Necessity by action type (\texttt{Type}: 82.0\%, \texttt{Select}: 79.7\%).}
\end{subfigure}
\\[0.5em]
\begin{subfigure}[b]{0.48\textwidth}
    \centering
    \includegraphics[width=\textwidth]{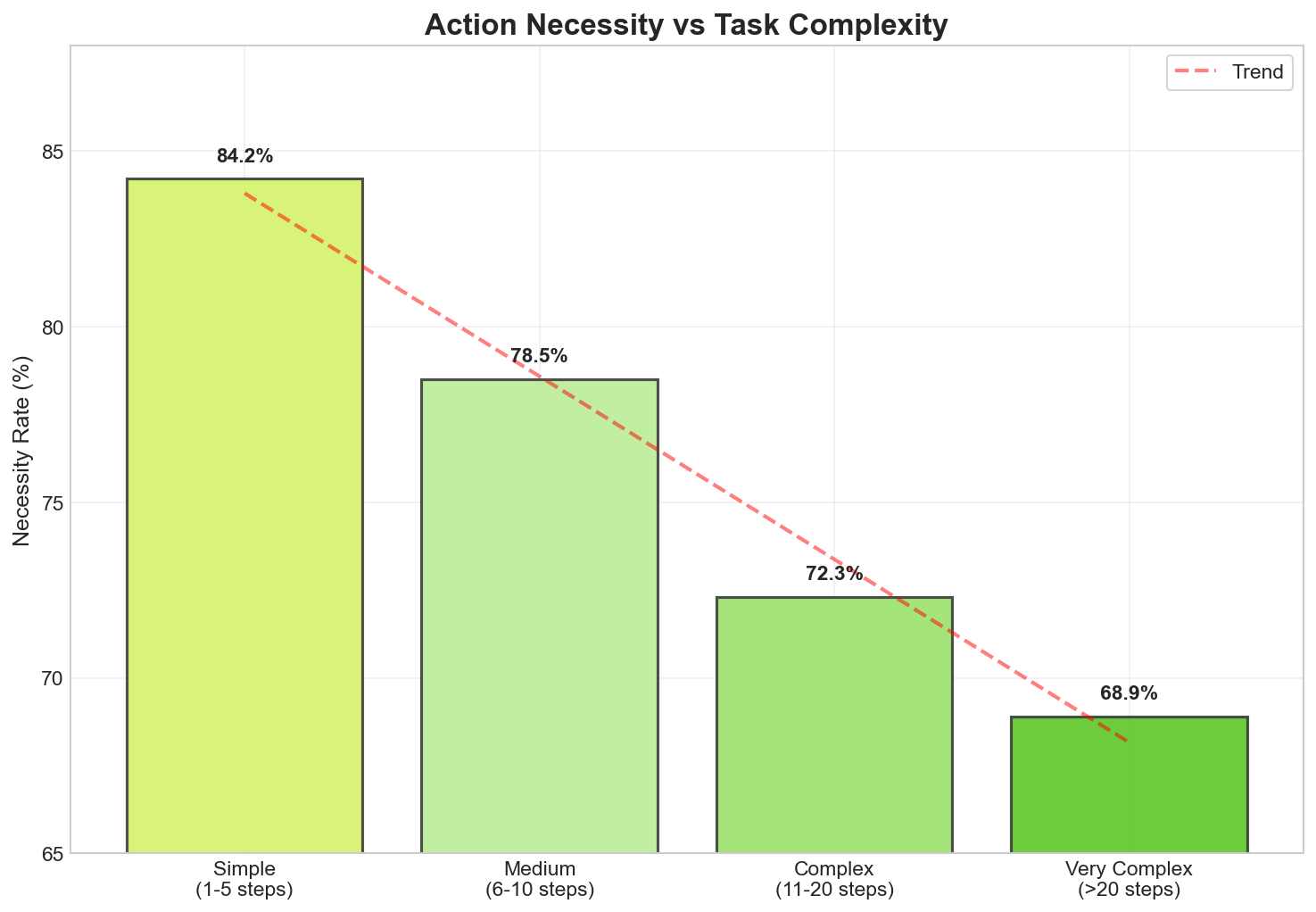}
    \caption{Necessity vs. task complexity (drops from 84.2\% to 68.9\%).}
\end{subfigure}
\hfill
\begin{subfigure}[b]{0.48\textwidth}
    \centering
    \includegraphics[width=\textwidth]{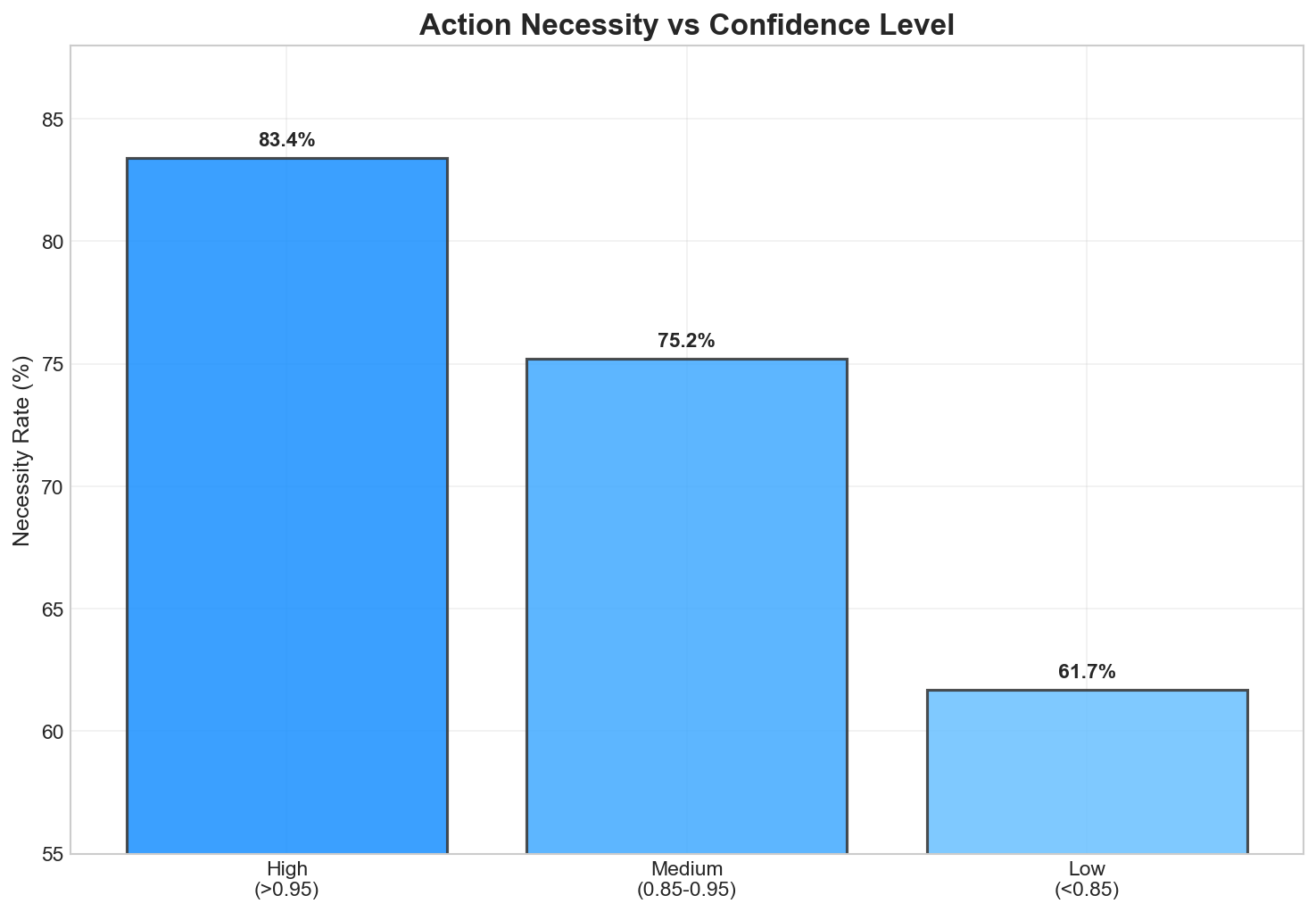}
    \caption{Necessity by action confidence ($r=0.67$, $p<0.001$).}
\end{subfigure}
\caption{Core necessity metrics across four dimensions: overall, action type, complexity, and confidence.}
\label{fig:appendix_necessity}
\end{figure}
\begin{figure}[ht]
\centering
\begin{subfigure}[b]{0.48\textwidth}
    \includegraphics[width=\textwidth]{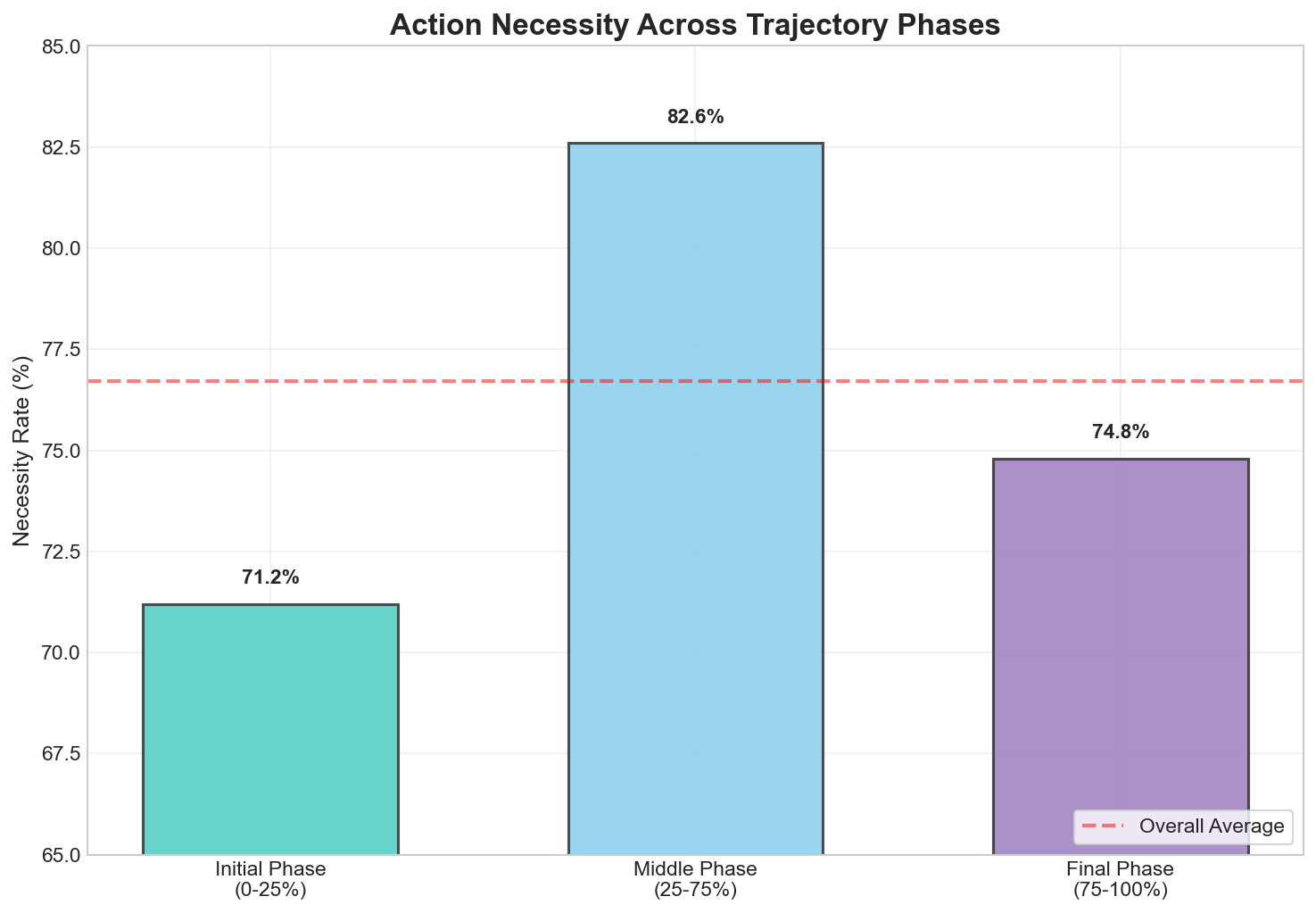}
    \caption{Temporal necessity patterns (peak in middle phase, 82.6\%).}
\end{subfigure}
\hfill
\begin{subfigure}[b]{0.48\textwidth}
    \includegraphics[width=\textwidth]{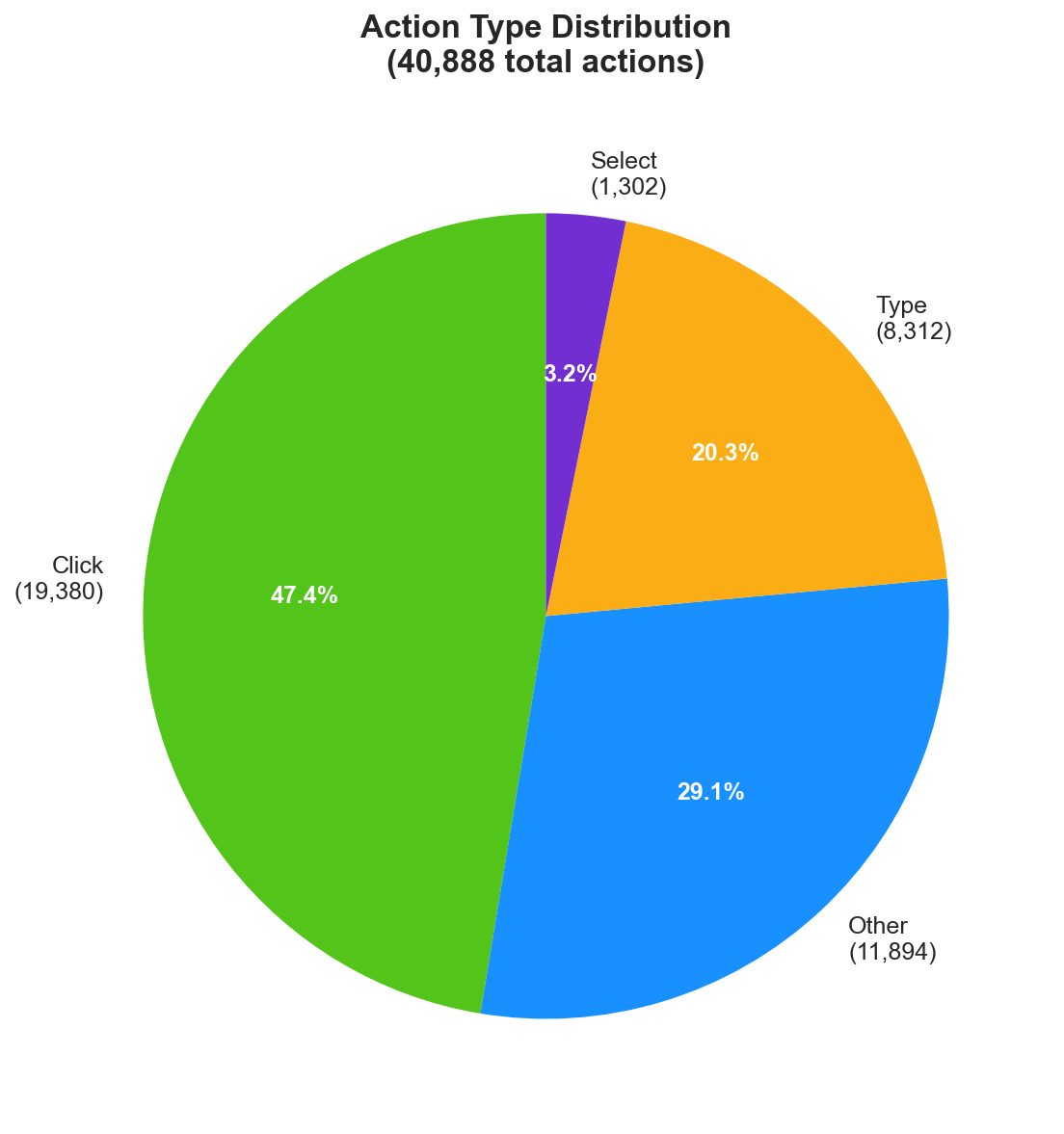}
    \caption{Action type distribution (click dominates at 47.4\%).}
\end{subfigure}
\caption{Supplementary behavioral patterns: (a) necessity varies across trajectory phases, with peak efficiency in the middle; (b) action type distribution reveals clicks dominate overall interactions.}
\label{fig:appendix_behavioral_efficiency}
\end{figure}

\subsection{Framework Performance}

\begin{figure}[ht]
\centering
\includegraphics[width=0.8\textwidth]{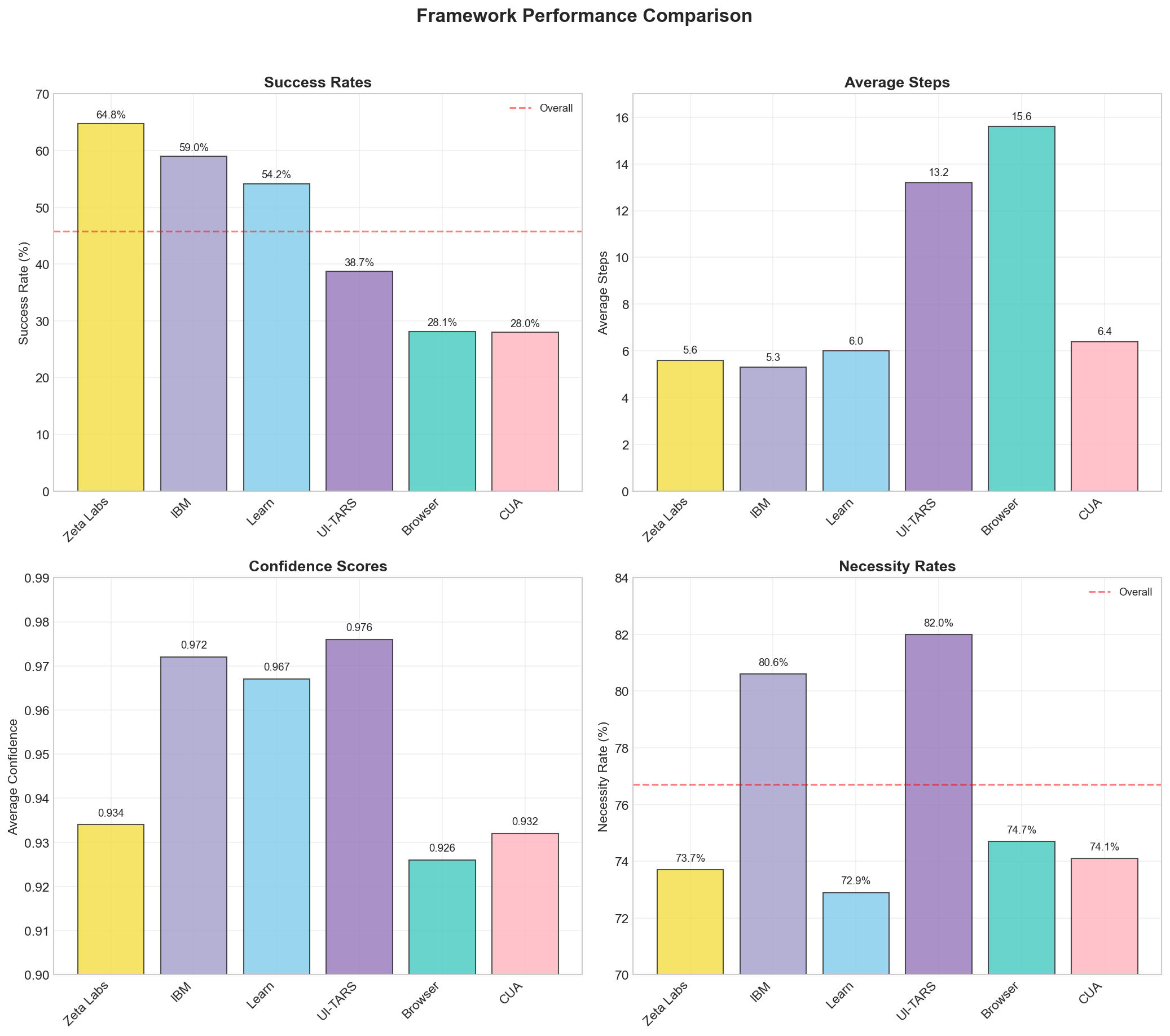}
\caption{Comprehensive framework performance matrix comparing six web agent frameworks across success rate, average steps, confidence scores, and necessity rates.}
\label{fig:appendix_framework}
\end{figure}

\subsection{Dataset Characteristics}

\begin{figure}[ht]
\centering
\includegraphics[width=0.55\textwidth]{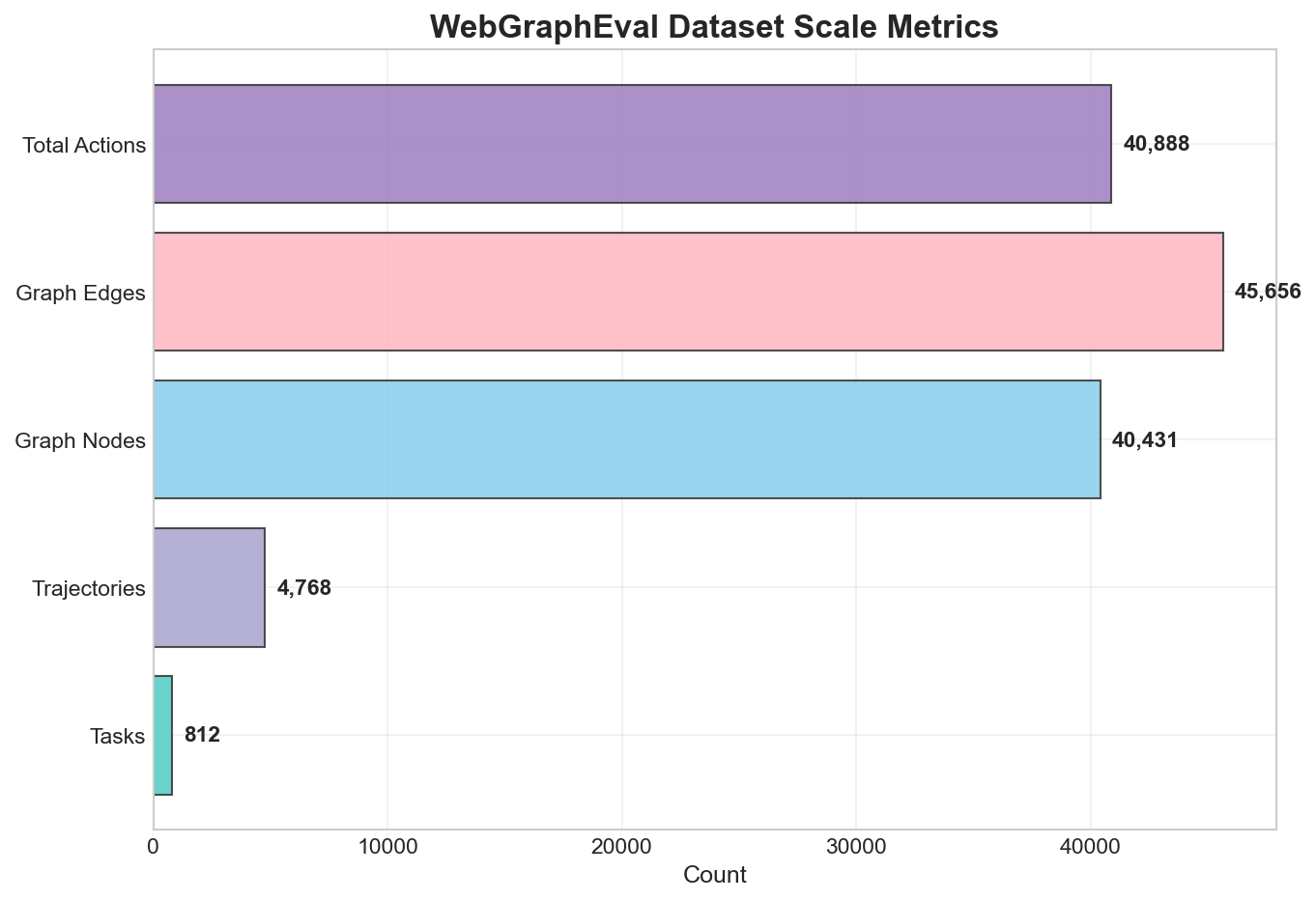}
\caption{Dataset scale: 812 tasks, 4,768 trajectories, 40,431 nodes, 45,656 edges, and 40,888 actions.}
\label{fig:appendix_dataset_scale}
\end{figure}

\begin{figure}[ht]
\centering
\begin{subfigure}[b]{0.48\textwidth}
    \includegraphics[width=\textwidth]{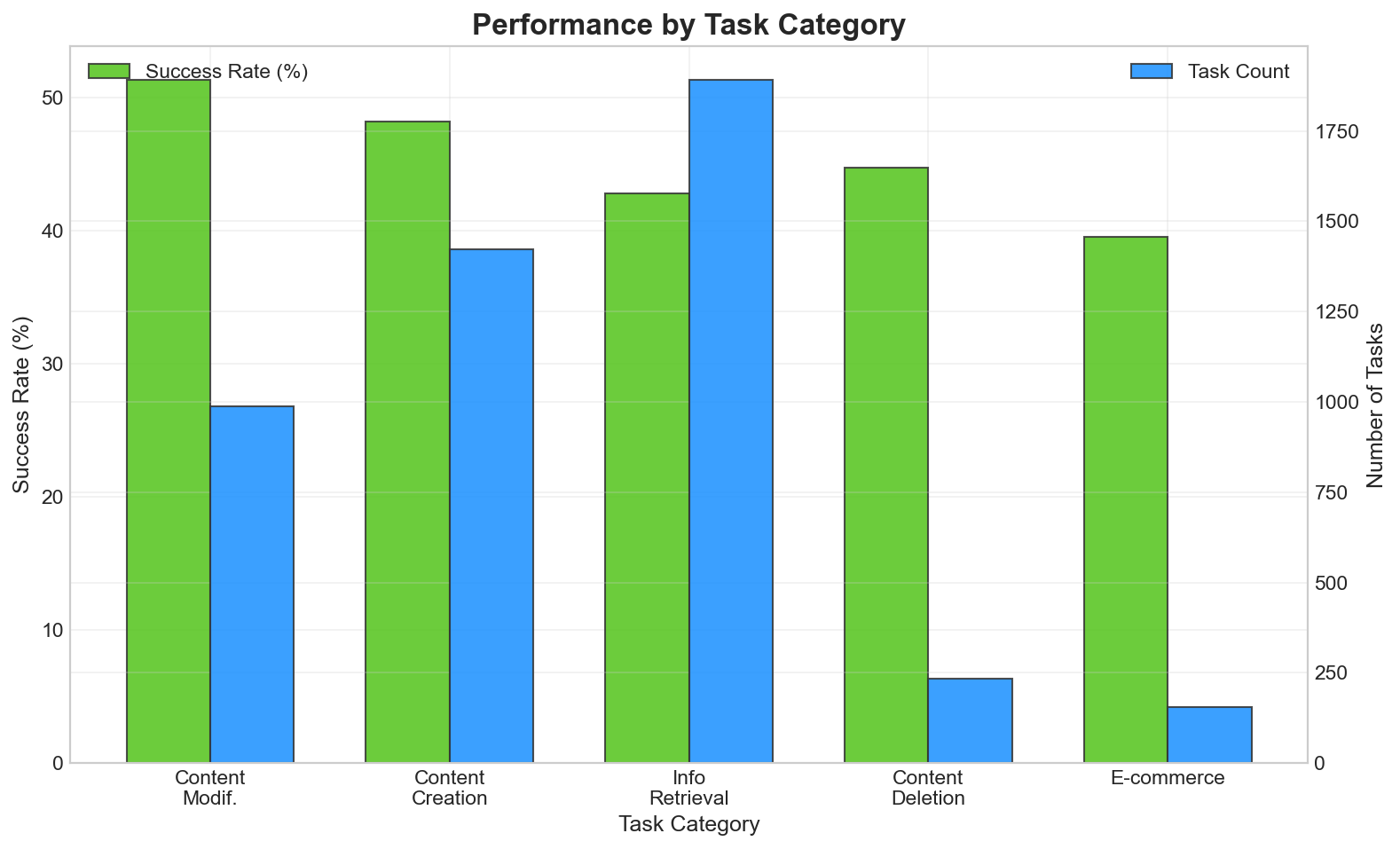}
    \caption{Performance by task category (content modification highest at 51.3\%).}
\end{subfigure}
\hfill
\begin{subfigure}[b]{0.48\textwidth}
    \includegraphics[width=\textwidth]{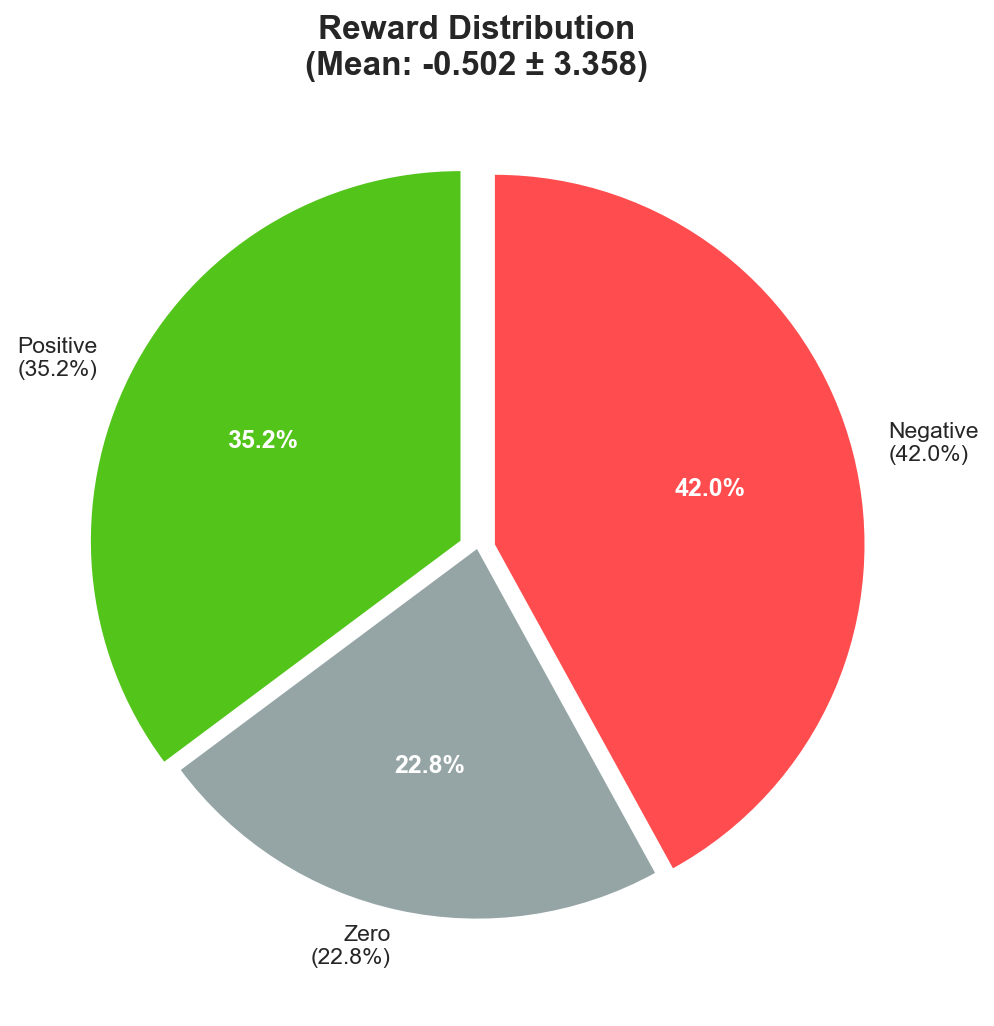}
    \caption{Reward distribution (35.2\% positive, 42.0\% negative).}
\end{subfigure}
\caption{Task-level analysis by category and reward distribution.}
\label{fig:appendix_task}
\end{figure}

\begin{figure}[ht]
\centering
\begin{subfigure}[b]{0.48\textwidth}
    \includegraphics[width=\textwidth]{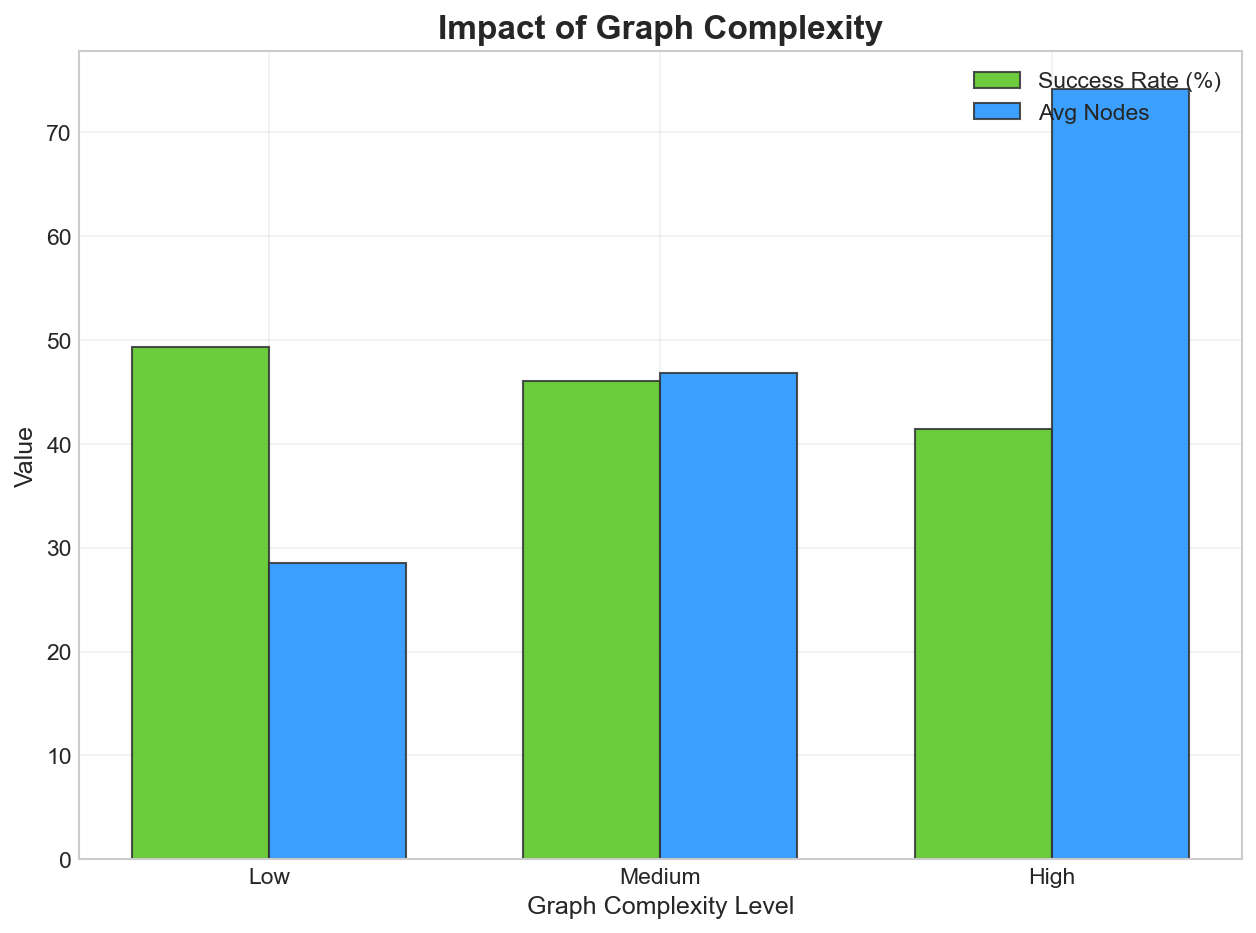}
    \caption{Impact of graph complexity (success decreases 7.9\% from low to high).}
\end{subfigure}
\hfill
\begin{subfigure}[b]{0.48\textwidth}
    \includegraphics[width=\textwidth]{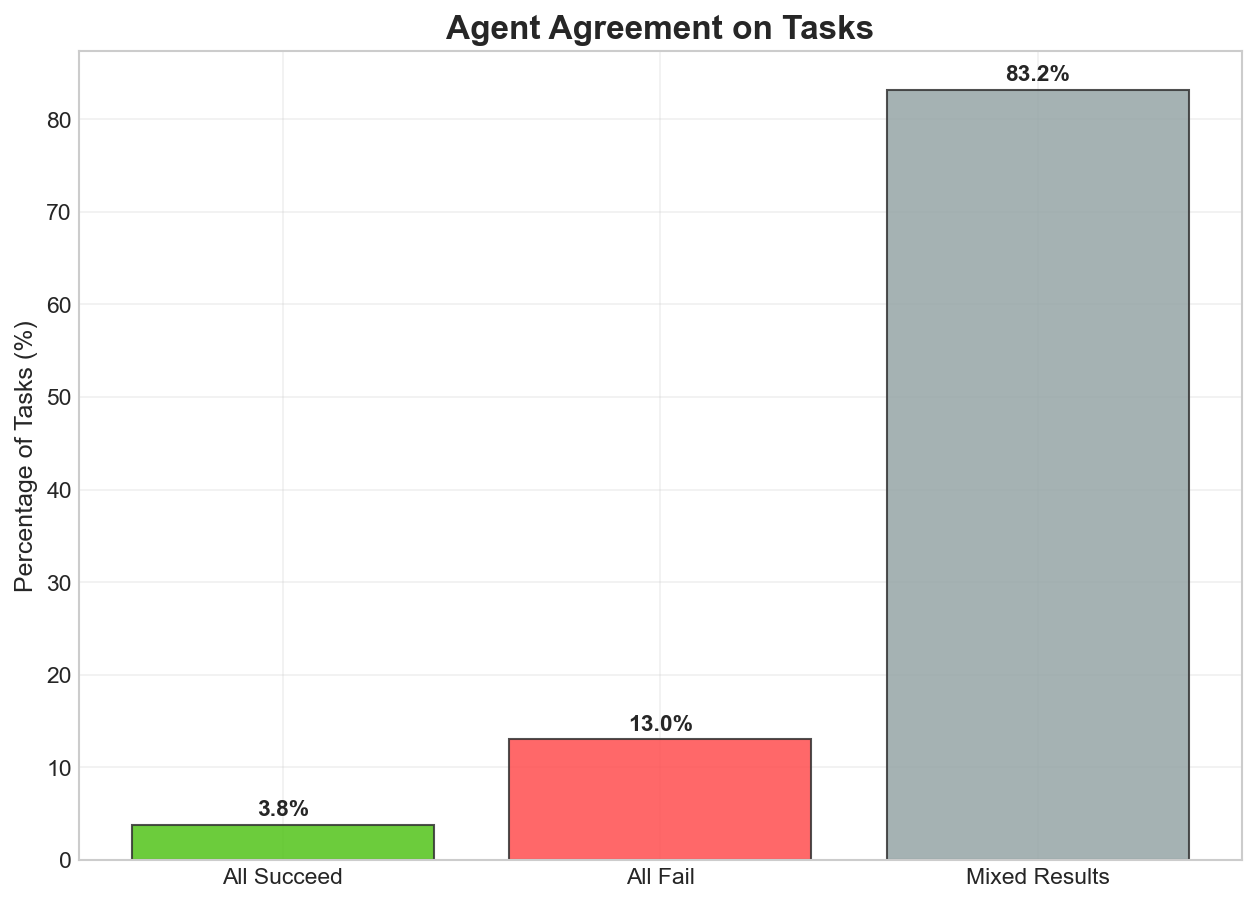}
    \caption{Agent agreement (3.8\% all succeed, 13.0\% all fail, 83.2\% mixed).}
\end{subfigure}
\caption{Complexity and agent agreement patterns.}
\label{fig:appendix_complexity}
\end{figure}

\begin{figure}[ht]
\centering
\begin{subfigure}[b]{0.48\textwidth}
    \includegraphics[width=\textwidth]{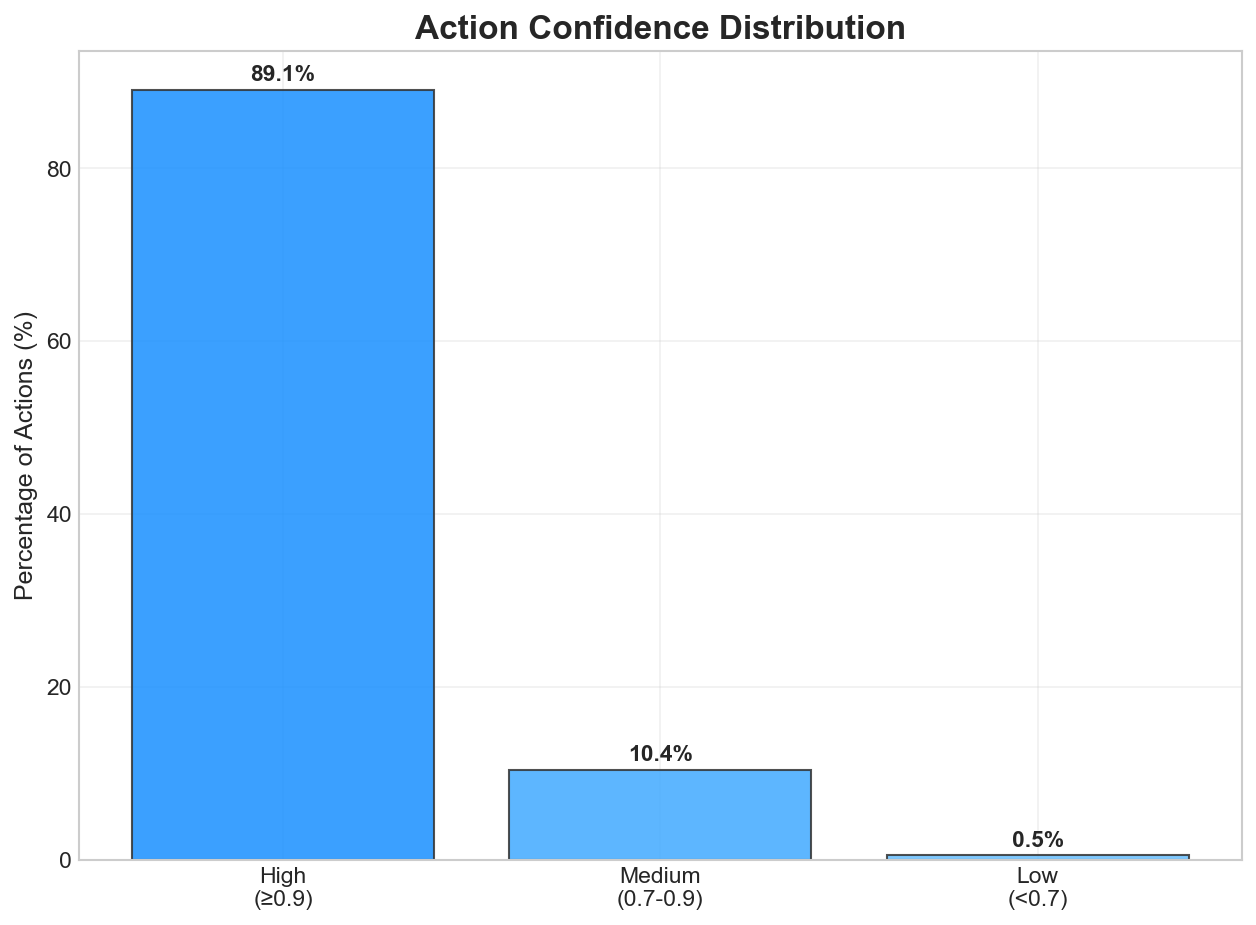}
    \caption{Action confidence distribution (89.1\% high-confidence).}
\end{subfigure}
\hfill
\begin{subfigure}[b]{0.48\textwidth}
    \includegraphics[width=\textwidth]{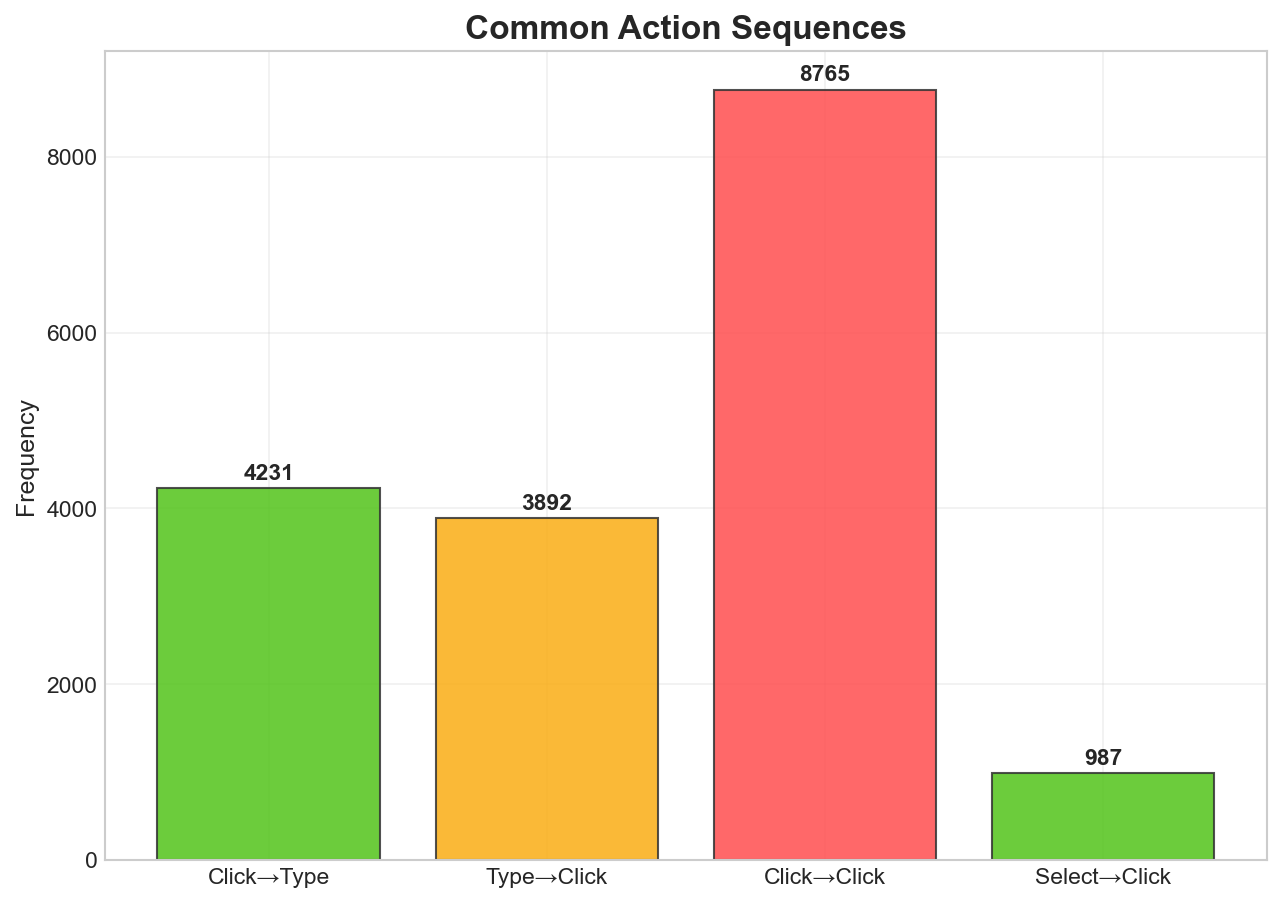}
    \caption{Frequent action sequences (Click→Click, Click→Type).}
\end{subfigure}
\caption{Confidence and sequence-level patterns.}
\label{fig:appendix_patterns}
\end{figure}

\begin{figure}[ht]
\centering
\includegraphics[width=0.65\textwidth]{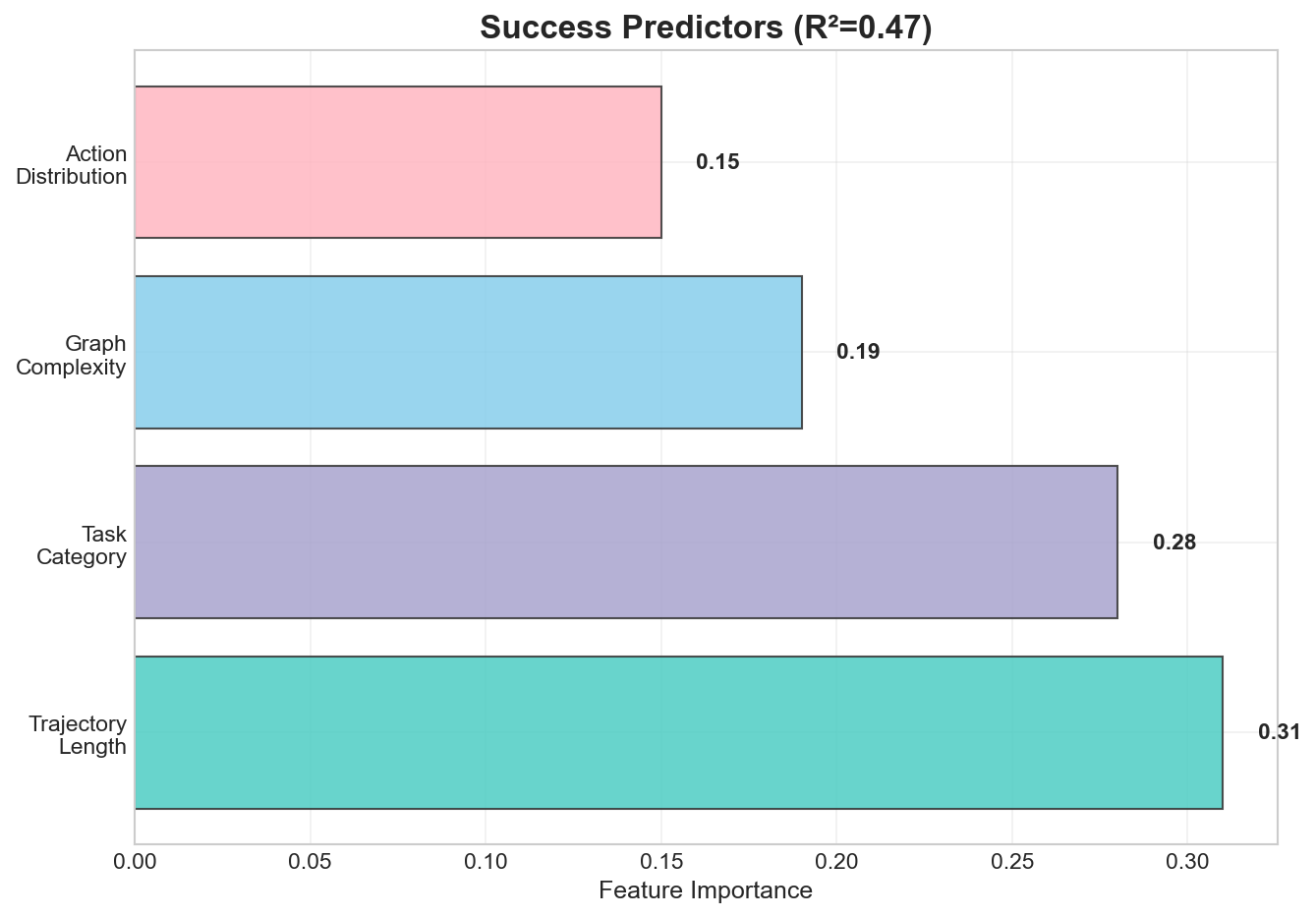}
\caption{Success predictors: trajectory length (0.31), task category (0.28), graph complexity (0.19), action distribution (0.15). Overall R²=0.47.}
\label{fig:appendix_predictors}
\end{figure}

\begin{table}[h]
\centering
\begin{minipage}[t]{0.45\textwidth}
\centering
\caption{Overall Graph Dataset Statistics}
\label{tab:dataset-stats}
\begin{tabular}{@{}lc@{}}
\toprule
\textbf{Metric} & \textbf{Value} \\
\midrule
Total Graph Nodes & 40,431 \\
Total Graph Edges & 45,656 \\
Average Nodes per Task & 49.79 $\pm$ 21.83 \\
Average Edges per Task & 56.23 $\pm$ 22.86 \\
Average Steps per Trajectory & 8.58 \\
\bottomrule
\end{tabular}
\end{minipage}
\hfill
\begin{minipage}[t]{0.50\textwidth}
\centering
\caption{Framework-level performance with mean $\pm$ standard deviation over 3 evaluation runs.}
\label{tab:framework-performance-std}
\begin{tabular}{@{}lc@{}}
\toprule
\textbf{Framework} & \textbf{Success Rate (mean ± std)} \\
\midrule
\textbf{Zeta Labs Jace.AI} \cite{jaceai2025} & 64.86\% ± 0.43 \\
\textbf{IBM CUGA} \cite{ibm-cuga-2024} & 59.70\% ± 0.14 \\
\textbf{Learn by Interact} \cite{learnbyinteract} & 53.74\% ± 0.43 \\
\textbf{UI-TARS} \cite{uitars} & 38.88\% ± 0.35 \\
\textbf{OpenAI-CUA} \cite{openai-cua-2025} & 28.55\% ± 0.07 \\
\textbf{BrowserUse} \cite{browseruse2025} & 27.66\% ± 0.15 \\
\bottomrule
\end{tabular}%
\end{minipage}
\end{table}

\begin{table*}
\centering
\caption{Comparison of graph-based approaches across domains. Prior work applies graphs for knowledge retrieval or causal reasoning, while WebGraphEval uses graphs to aggregate trajectories for evaluation.}
\label{tab:graph-approaches}
\resizebox{\textwidth}{!}{%
\begin{tabular}{@{}lllllll@{}}
\toprule
\textbf{Method} & \textbf{Graph Type} & \textbf{Node Type} & \textbf{Edge Type} & \textbf{Multi-hop} & \textbf{Method} & \textbf{Domain} \\
\midrule
GraphRAG \citep{edge2024local} & Knowledge Graph & Entities & Semantic & \ding{51} & Traversal & Document QA \\
LightRAG \citep{guo2024lightrag} & Sparse Graph & Entities & Semantic & \ding{51} & Traversal & Real-time QA \\
Causal GraphRAG \citep{haque2025graphrag} & Causal Graph & Events & Causal & \ding{51} & Path-finding & News Analysis \\
Graph-R1 \citep{luo2025graph} & Hypergraph & Entities & Hyperedges & \ding{51} & Interaction & Complex QA \\
\midrule
\rowcolor{highlight}
\textbf{WebGraphEval (Ours)} & Directed Graph & Canonical Actions & Transitions & \ding{51} & Trajectory Aggregation & Web Agent Evaluation \\
\bottomrule
\end{tabular}%
}
\end{table*}

\begin{figure}[t]
\centering
\begin{tcolorbox}[colback=blue!5!white, colframe=blue!75!black, 
title=Trajectory Success Evaluation, boxrule=0.3mm, width=\textwidth, arc=3mm, auto outer arc=true]

\textbf{[System Prompt]}\\[3mm]

You are an expert evaluator for web agent trajectories. Your task is to determine if a web agent successfully completed a given task based on its actions and final response.

\medskip

\textbf{EVALUATION CRITERIA:}
\begin{enumerate}
    \item Analyze the agent's actions, URLs visited, and final message
    \item Check if the agent fulfilled the task intent
    \item If reference answers are provided, verify they appear in the trajectory
    \item Consider both the journey (actions/URLs) and destination (final message)
    \item Be strict but fair in your assessment
\end{enumerate}

\medskip

\textbf{RESPONSE FORMAT:}\\
Respond with exactly \texttt{"SUCCESS"} or \texttt{"FAILURE"} on the first line, followed by a brief explanation on the next line.

\medskip
\rule{\linewidth}{0.2pt} 
\medskip

\textbf{[User Prompt]}\\[3mm]

Task Intent: \{intent\}

\medskip

Reference Answers:
\begin{itemize}
    \item Exact match expected: ``\{exact\_match\}''
    \item Must include all of: ``\{must\_include\_items\}''
    \item Fuzzy match acceptable: ``\{fuzzy\_match\_items\}''
\end{itemize}

\medskip

Final Message: \\
``\{agent final reply\}''

\medskip

Action Sequence: \\
\{index\}. \{action\} (at \{url\}) \\
\dots

\medskip

\textbf{EVALUATION CRITERIA (\{with / without\} Reference Answer):}
\begin{enumerate}
    \item Does the final message contain the correct answer that matches the reference?
    \item If the answer is not in the final message, check if it appears in the actions or URLs
    \item The reference answer \textbf{MUST} be found somewhere in the trajectory for success
\end{enumerate}

\medskip

\textbf{IMPORTANT:} The provided reference answers are the ground truth. Success requires finding these specific answers.

\medskip

Respond with: \texttt{SUCCESS} or \texttt{FAILURE}, followed by a brief explanation.

\end{tcolorbox}
\caption{Trajectory Success Evaluation prompt}
\label{fig:prompt-evaluation}
\end{figure}

\begin{figure}[t]
\centering
\begin{tcolorbox}[colback=green!5!white, colframe=green!75!black, 
title=System Prompt: Action Conversion and Necessity Annotation, 
boxrule=0.3mm, width=\textwidth, arc=3mm, auto outer arc=true]

You are an expert at converting natural language web action descriptions into standardized function calls.

\medskip
\textbf{Available Functions:}
\begin{itemize}
  \item \texttt{click(text: string, element?: string)} – Click on an element
  \item \texttt{type(text: string, element?: string)} – Type text into an input field
  \item \texttt{scroll(direction: "up"|"down", amount?: number)} – Scroll the page
  \item \texttt{select(value: string, element?: string)} – Select from dropdown
  \item \texttt{hover(text: string, element?: string)} – Hover over element
  \item \texttt{wait(seconds: number)} – Wait for specified time
  \item \texttt{goto(url: string)} – Navigate to URL
  \item \texttt{back()} – Go back in browser history
  \item \texttt{refresh()} – Refresh the page
\end{itemize}

\medskip
\textbf{Instructions:}
\begin{enumerate}
  \item Analyze the input action description carefully
  \item Extract the key intent and parameters
  \item Map to the most appropriate function from the available list
  \item Use named parameters format: \texttt{functionName(param1="value1", param2="value2")}
  \item If multiple interpretations are possible, choose the most likely one
  \item Maintain high confidence for clear matches, lower for ambiguous cases
  \item If no function matches well, return confidence $< 0.5$
  \item ALSO decide whether the action is necessary for accomplishing the task (boolean \texttt{necessary})
  \item If the action depends on earlier steps, include a nested \texttt{pre} field: \{id:"<step id>", pre:\{...\}\}
  \item If the input describes multiple discrete actions, split them and output a JSON array
\end{enumerate}

\medskip
\textbf{Output Format (JSON):}
\begin{verbatim}
{
  "functionName": "click",
  "parameters": ["Submit", "button"],
  "namedParameters": {"text": "Submit", "element": "button"},
  "confidence": 0.95,
  "necessary": true,
  "pre": { "id": "step 2" },
  "reasoning": "Clear click action on a button element"
}
\end{verbatim}

\textbf{Important:}
\begin{itemize}
  \item Always include both \texttt{"parameters"} (array) and \texttt{"namedParameters"} (object)
  \item Always include \texttt{"necessary"}
  \item Always include \texttt{"pre"} (null if no dependency)
\end{itemize}

Always output valid JSON. Be concise but accurate.

\end{tcolorbox}
\caption{System prompt for action conversion and necessity annotation.}
\label{fig:prompt-action-conversion-system}
\end{figure}

\begin{figure}[t]
\centering
\begin{tcolorbox}[colback=blue!5!white, colframe=blue!75!black, 
title=User Prompt: Action Conversion and Necessity Annotation, 
boxrule=0.3mm, width=\textwidth, arc=3mm, auto outer arc=true]

You are given a natural language description that may contain multiple action sentences. First split the text into individual sentences by period (.), exclamation (!), or question mark (?). For each non-empty sentence, produce a standardized function call. Output a JSON array of conversion objects, one per sentence.

\medskip
Description: \\
``\{action\}''

\medskip
Task Context: \{task\_description\}

\medskip
Previous Steps:
\begin{itemize}
  \item [step 0] \{previous\_action\_0\}
  \item [step 1] \{previous\_action\_1\}
  \item $\ldots$
\end{itemize}

\end{tcolorbox}
\caption{User prompt for action conversion and necessity annotation.}
\label{fig:prompt-action-conversion-user}
\end{figure}